\documentclass[10pt,twocolumn,letterpaper]{article}
\usepackage{cvpr}              

\usepackage{graphics} 
\usepackage{epsfig} 
\usepackage{mathptmx} 
\usepackage{times} 
\usepackage{amsfonts} 
\usepackage{amssymb}  
\usepackage{amsmath} 

\usepackage{graphicx}
\usepackage{helvet}

\usepackage{url}
\usepackage{rotating}
\usepackage{xspace}
\usepackage{booktabs}       
\usepackage{nccmath}
\usepackage{nicefrac}       
\usepackage{microtype}      
\usepackage{comment}		
\usepackage{adjustbox}		
\usepackage{tabu}
\usepackage{calc}			
\usepackage{wrapfig}
\usepackage[pagebackref,breaklinks,colorlinks,bookmarks=false]{hyperref}

\usepackage{multirow}
\usepackage{float}
\usepackage[hang,flushmargin]{footmisc}
\usepackage[neverdecrease]{paralist}
\usepackage{caption}
\usepackage{algorithm}
\usepackage{algorithmicx}
\usepackage{algpseudocode}

\usepackage{calc}
\usepackage{color}
\usepackage{colortbl}
\usepackage{animate}

\usepackage[capitalize]{cleveref}
\crefname{section}{Sec.}{Secs.}
\Crefname{section}{Section}{Sections}
\Crefname{table}{Table}{Tables}
\crefname{table}{Tab.}{Tabs.}

\newcommand{\figref}[1]{Fig.~\ref{#1}}
\newcommand{\tabref}[1]{Table~\ref{#1}}
\newcommand{\eqnref}[1]{Eq.~(\ref{#1})}
\newcommand{\secref}[1]{Sec.~\ref{#1}}

\usepackage[mathcal]{eucal}

\usepackage{ntheorem}

\def\cvprPaperID{*****} 
\def\confName{CVPR}
\def\confYear{2023}

\begin{document}

\title{Rotation Matters: Generalized Monocular 3D Object Detection for Various Camera Systems}
\author{Sungho Moon, Jinwoo Bae, and Sunghoon Im\\
Department of Electrical Engineering \& Computer Science, DGIST, Daegu, Korea\\
{\tt\small \{byeol3325, sjg02122, sunghoonim\}@dgist.ac.kr}
}
\affiliation{Department of Electrical Engineering \& Computer Science, DGIST, Daegu, Korea}

\maketitle

\begin{abstract}
Research on monocular 3D object detection is being actively studied, and as a result, performance has been steadily improving. 
However, 3D object detection performance is significantly reduced when applied to a camera system different from the system used to capture the training datasets. 
For example, a 3D detector trained on datasets from a passenger car mostly fails to regress accurate 3D bounding boxes for a camera mounted on a bus.
In this paper, we conduct extensive experiments to analyze the factors that cause performance degradation.
We find that changing the camera pose, especially camera orientation, relative to the road plane caused performance degradation.
In addition, we propose a generalized 3D object detection method that can be universally applied to various camera systems. We newly design a compensation module that corrects the estimated 3D bounding box location and heading direction.
The proposed module can be applied to most of the recent 3D object detection networks.
It increases AP3D score (KITTI moderate, IoU $> 70\%$) about 6-to-10-times above the baselines without additional training. 
Both quantitative and qualitative results show the effectiveness of the proposed method. 




\end{abstract}
\section{Introduction}


3D object detection is the task of estimating the 3D position and orientation of multiple objects in a scene.
It plays an important role in various visual perception applications such as autonomous driving systems and robot bin picking.
To detect objects in 3D space, conventional methods use various sensors including single cameras, stereo cameras, LiDAR, RADAR or a fusion of multiple sensors \cite{pang2020clocs,fan2021rangedet,sun2021rsn,meyer2019automotive,meyer2019deep,nabati2021centerfusion}.
In particular, recently, single-camera-based 3D object detection, or so-called monocular 3D object detection~\cite{gu2022homography,zhang2021objects,ma2021delving} has attracted increasing interest because a single camera system is cost-effective, light-weight and easily mountable.



Monocular 3D object detection is a highly challenging problem because depth information is typically lacking ~\cite{chen2016monocular}.
Recent methods~\cite{lian2021geometry}, \cite{zhang2021objects}, \cite{ma2021delving} decouple the 3D bounding box regression problem into several progressive sub-tasks such as estimating the object 3D center, 3D bounding box size, and 3D heading direction.
These methods disassemble the elements of the 3D bounding box and impose a regression loss for each parameter. This helps to train the entire network effectively and to analyze the contribution of each component. These works currently achieve state-of-the-art performance. 
However, these methods generally do not apply to camera systems mounted on other vehicles (\textit{e.g.} passenger cars, trucks, and buses) even when the same camera model is used. The camera position and orientation are uniquely set based on the vehicle size and platform.
Changes in camera poses drastically degrade the 3D object detection performance.

In this paper, we investigate the root causes of performance degradation. 
To do so, we synthetically generate various images and their corresponding 3D bounding box labels by changing either rotation or translation or both. 
Through extensive experiments, we observe that state-of-the-art monocular 3D object detectors~\cite{qin2019monogrnet,liu2020smoke,ma2021delving,zhang2021objects} produce about $1\%$ $AP_{3D}$ score (KITTI moderate, IoU $> 70\%$) given images captured from different orientations.
Changing the camera orientation in a roll or pitch axis drastically degrades the 3D object detection performance, while changing the camera position and camera orientation in the yaw axis had little effect.
This is because the 3D object detectors have never been trained to regress the 3D heading direction of objects in roll and pitch angles. 
The methods assume that the camera mounted on the vehicle has a fixed position and orientation with respect to the road plane.
They parameterize the 3D heading direction of the vehicles as a single value for yaw-angle, instead of estimating all rotation parameters (roll, pitch, and yaw).

To tackle this issue, we propose a 3D heading compensation module, which is a simple yet effective algorithm for a generalized solution. 
It corrects the estimated object 3D head direction from conventional 3D object detectors, so additional training datasets and training steps are not required. We only need the relative camera orientation between the camera capturing the training data and the camera for the test data. 
We use the pre-calibrated camera extrinsic obtained in the manufacturing process. Extensive experiments with various datasets and ablation studies demonstrate the effectiveness of our method.
Our contributions can be summarized as follows:


\begin{itemize}
    \item We deeply analyze the individual prediction of the 3D object detector and figure out the factors that lead the performance degradation when the model is applied to other camera systems.
    \item We propose a generalized 3D object detection method that is trained on a specific camera setup but can be utilized in a variety of camera systems.
    \item The proposed method achieves a 6-to-10 times improvement compared to state-of-the-art methods without additional training. 
\end{itemize}

\begin{figure*}[t]
    \centering
        \includegraphics[width=12cm]{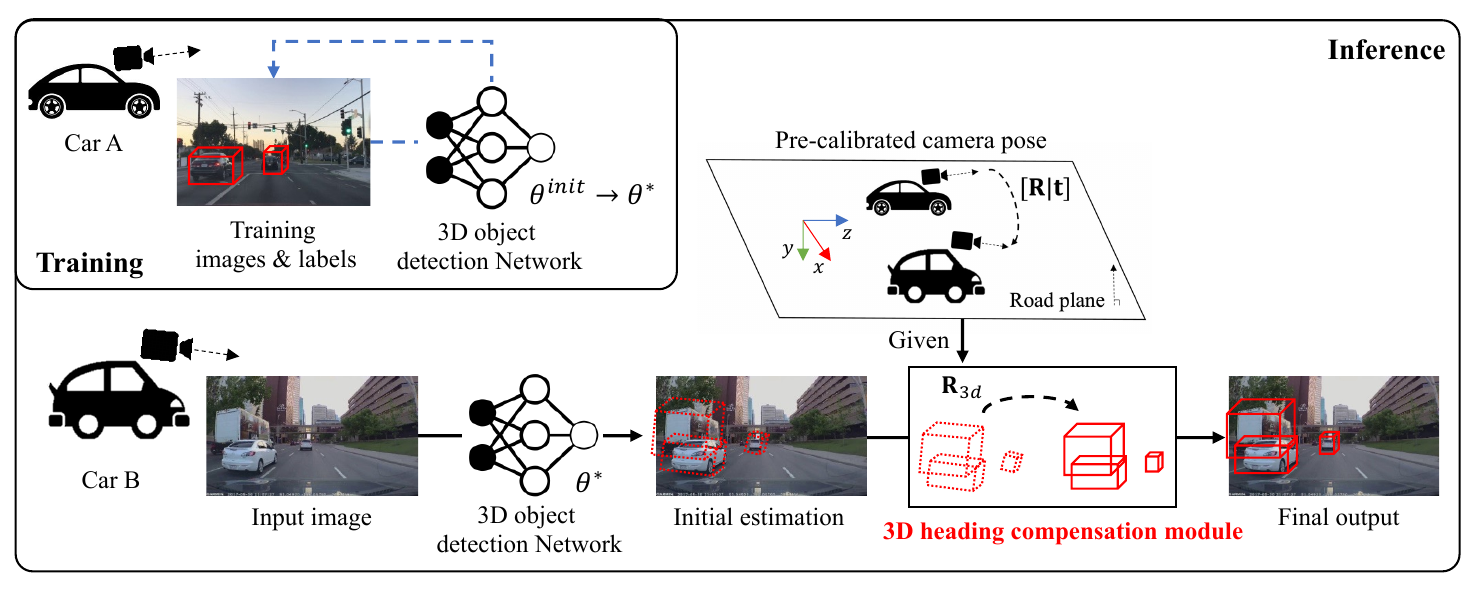}
    \vspace{-0.1cm}
    \caption{Overview of the proposed 3D object detection pipeline. We utilize a pretrained 3D object detection network (\textit{e.g.} MonoGRNet~\cite{qin2019monogrnet}, SMOKE~\cite{liu2020smoke}, MonoDLE~\cite{ma2021delving}, and MonoFlex~\cite{zhang2021objects}). The proposed 3D heading compensation module rotate the heading direction of initial estimation in yaw-axis. We use precalibrated camera pose $[\mathbf{R}|\mathbf{t}]$ between a source camera (training set) and a target camera (test set). }
    \vspace{-2mm}
\label{overview}
\end{figure*}

\section{Related Works}

\textbf{Monocular 3D Object Detection} Monocular 3D object detection methods estimate the 3D bounding box from a single RGB image. Estimating 3D information from only 2D information is a challenging problem. Mono3D \cite{chen2016monocular} utilizes the prior knowledge of car shape to estimate the 3D bounding box. DeepMANTA \cite{chabot2017deep} and ROI-10D \cite{manhardt2019roi} uses a 3D CAD model of vehicles and estimates the vehicle 3D bounding box using a robust 2D/3D vehicle part matching.
These methods require expensive amounts of training data including car shapes or 3D CAD models and require heavy computational time.
Another research direction incorporates a 2D detection network with a depth estimation network for monocular 3D object detection \cite{xu2018multi,qin2019monogrnet}. SMOKE \cite{liu2020smoke} predicts 3D object detection by combining 3D projected keypoints with regressed 3D regression parameters in an end-to-end manner. MonoDLE \cite{ma2021delving} estimates a coarse center, then reduces the location error between the 2D box center and the projected 3D box center. MonoFLEX \cite{zhang2021objects} separates the truncated object and the edge of the feature map to minimize the object depth estimation error. 

\textbf{Robust Monocular 3D Estimation to Camera Pose Changes}
Recently, monocular 3D geometry estimation tasks, such as depth estimation~\cite{baradad2020height,zhao2021camera} and 3D object detection~\cite{zhou2021monocular,li2021exploring,lian2021geometry} suffer from generalization issues with camera pose changes. 
Some works \cite{baradad2020height,zhao2021camera} propose generalized monocular depth estimation methods. 
The former predict camera pose and estimate depth in the world coordinates. The latter points out the problem of unbalanced distribution of camera extrinsic in training data, and tackles the issue through geometry-aware data augmentation.

Using monocular 3d object detection, Ego-Net \cite{li2021exploring} estimates the pose of each object relative to the camera pose to improve detection performance. Another work \cite{rangesh2020ground} inputs additional information such as a ground plane database or camera calibration parameters to detect the particular object, and is robustly accurate regardless of the plane. MonoEF \cite{zhou2021monocular} proposes a robust algorithm even with a change in camera extrinsic by fusing a visual odometry method and monocular 3d object detection.
In addition to MonoEF \cite{zhou2021monocular}, several recent methods try to solve the problem of robustness in monocular 3D object detection~\cite{li2021exploring,lian2021geometry}. However, existing methods require additional training of the model using the generated images, so the change is limited, unlike a real environment.




\section{Method}



%

\subsection{Image synthesis with different camera poses}
\label{sec:datageneration}
\begin{figure*}[t]
    \centering
    \begin{tabular}{c@{\hspace{1mm}}c@{\hspace{1mm}}c@{\hspace{1mm}}}
         \rotatebox[origin=l]{90}{MonoDLE} \includegraphics[width=0.3\linewidth]{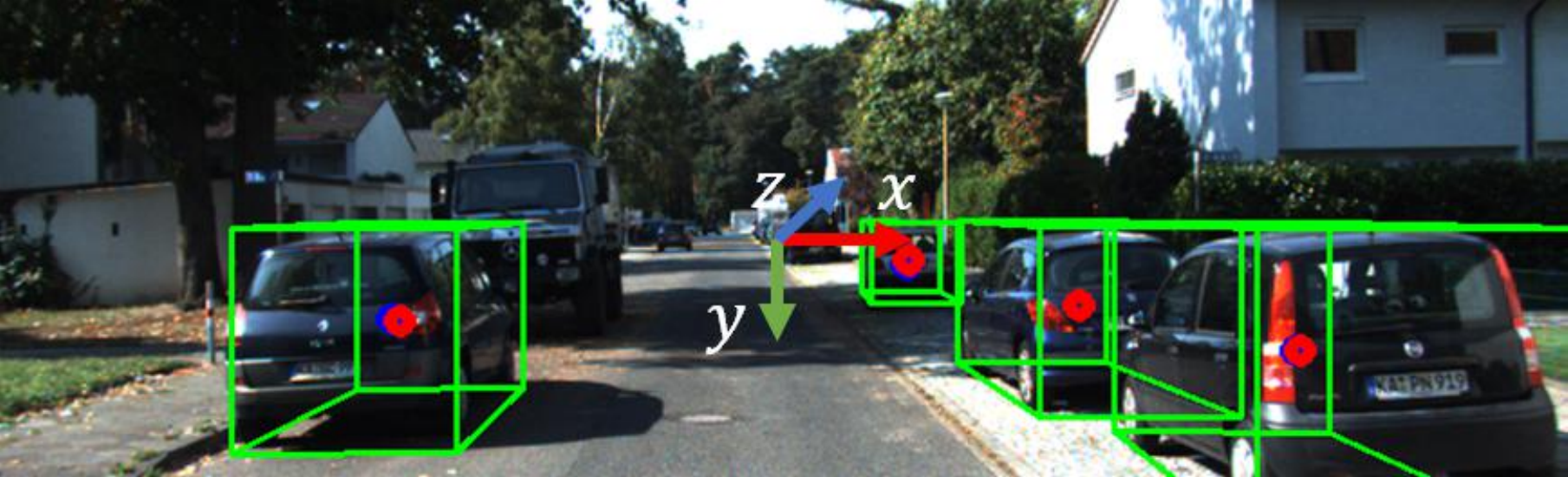}&
         \includegraphics[width=0.3\linewidth]{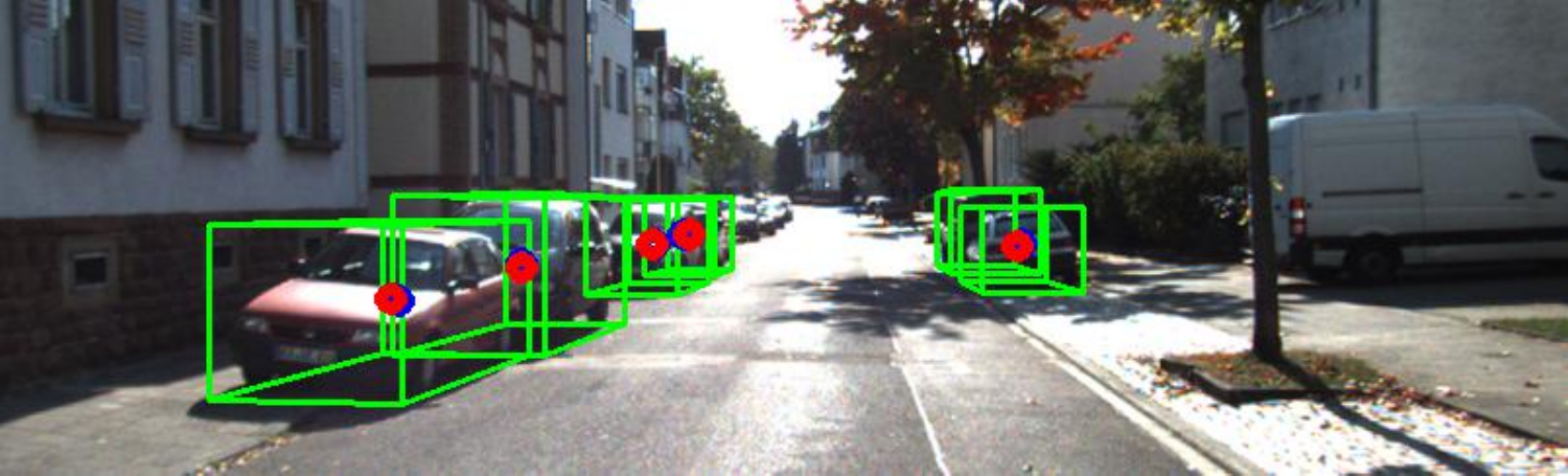}&
         \includegraphics[width=0.3\linewidth]{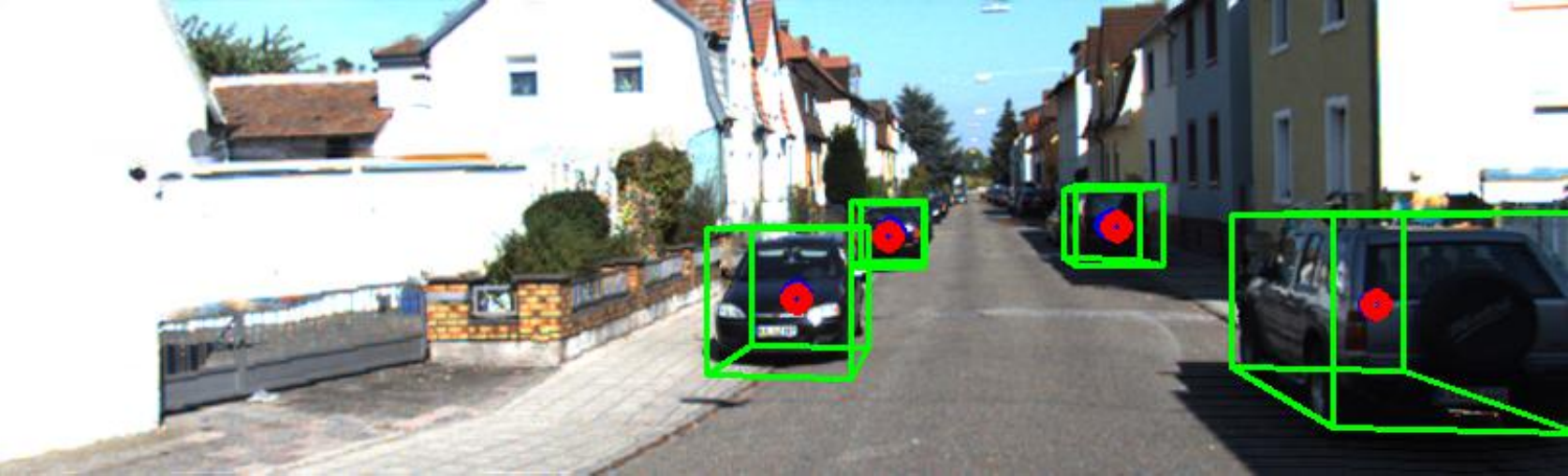} \\
         \rotatebox[origin=lc]{90}{      Ours} \includegraphics[width=0.3\linewidth]{figure/3D_results_vis/cropped/ori_crop.pdf}&
         \includegraphics[width=0.3\linewidth]{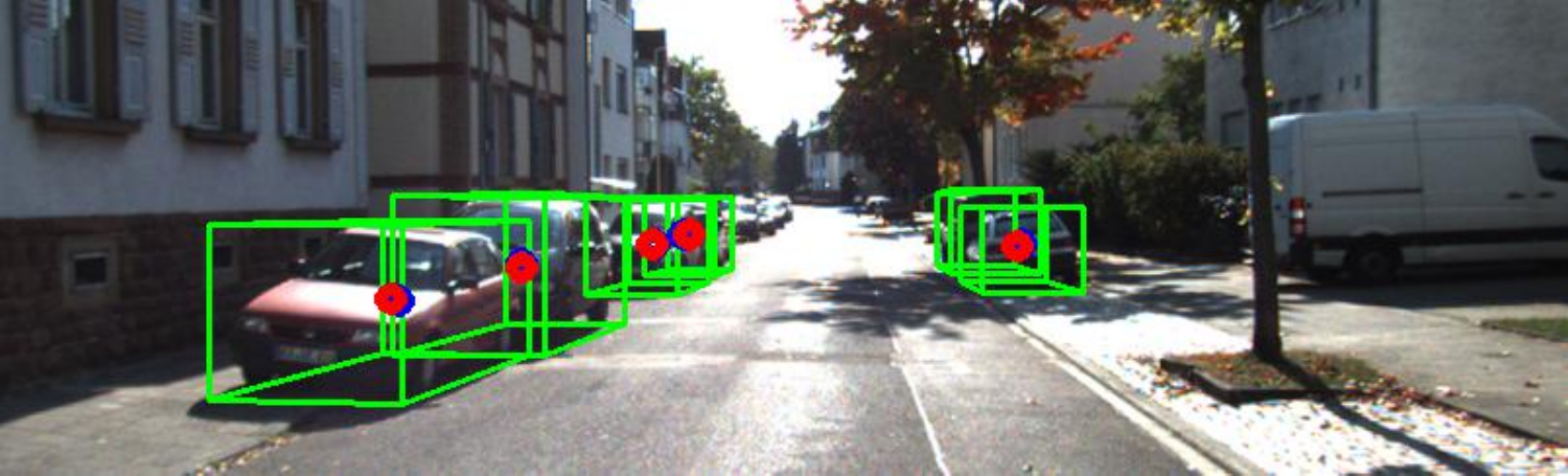}&
         \includegraphics[width=0.3\linewidth]{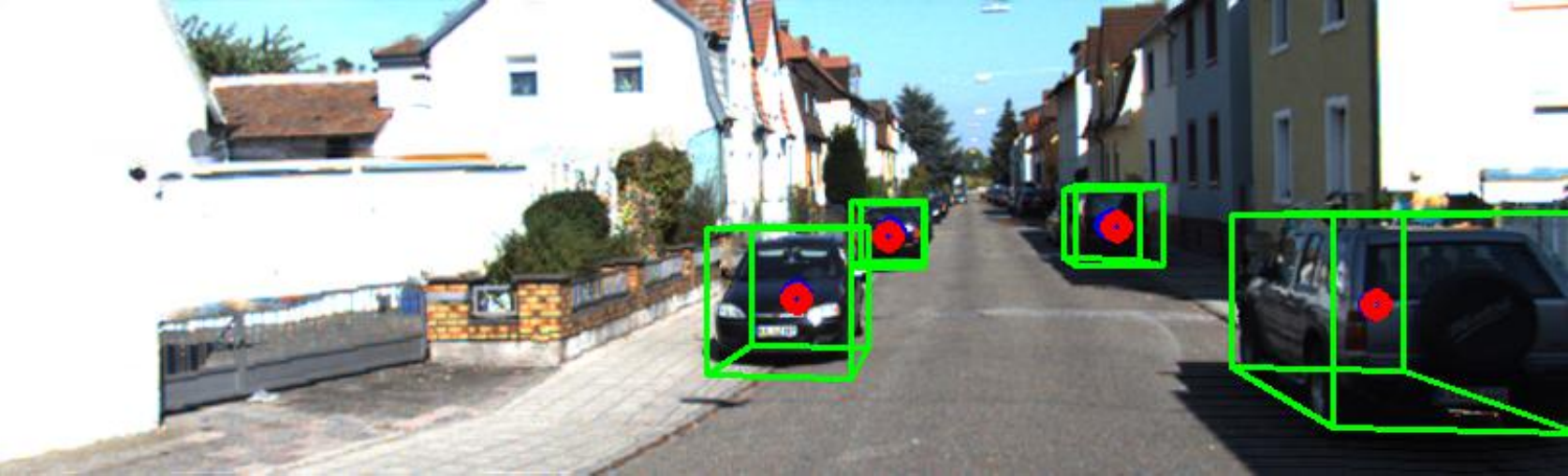} \\
         (a) Original & (b) $x$-translation & (c) $y$-translation\\
         \rotatebox[origin=l]{90}{MonoDLE} \includegraphics[width=0.3\linewidth]{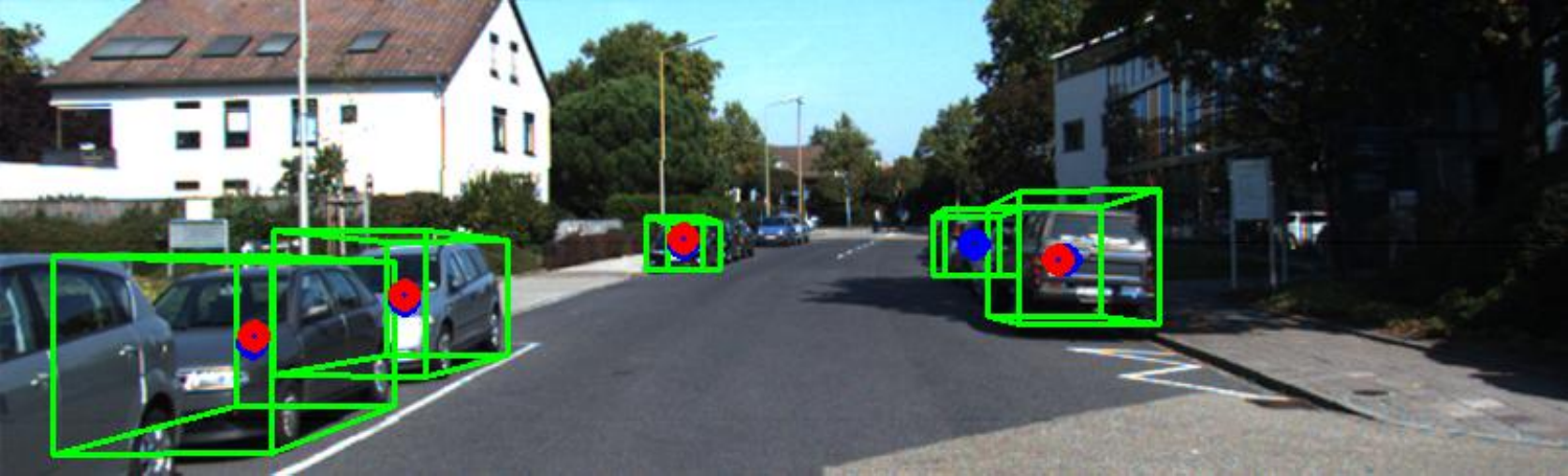}& 
         \includegraphics[width=0.3\linewidth]{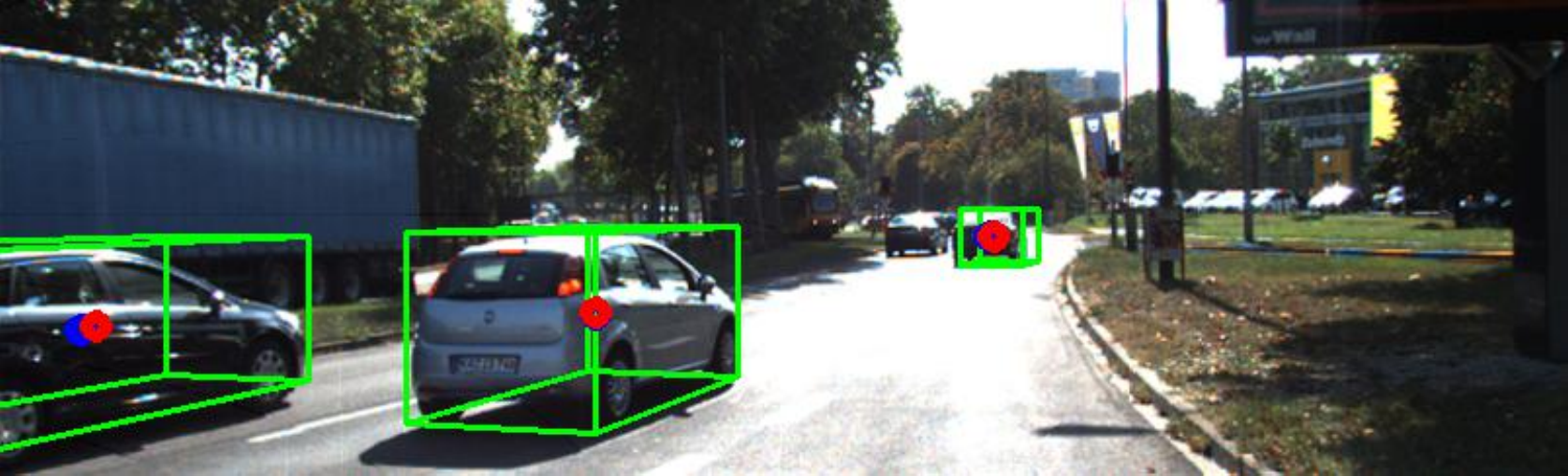}& 
         \includegraphics[width=0.3\linewidth]{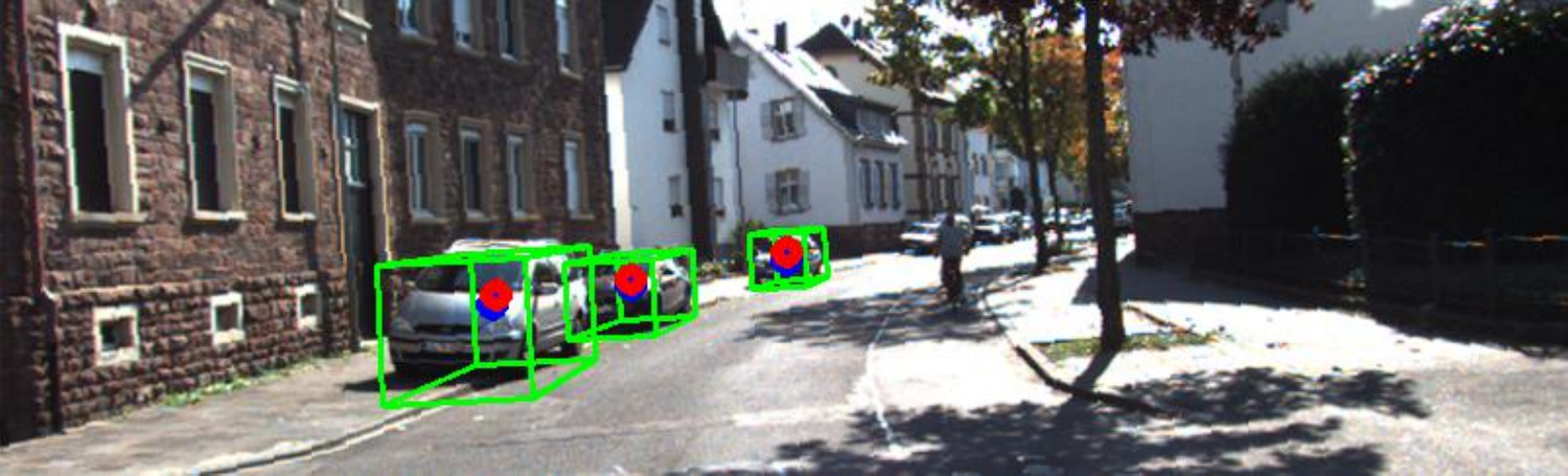}\\
         \rotatebox[origin=lc]{90}{Ours} \includegraphics[width=0.3\linewidth]{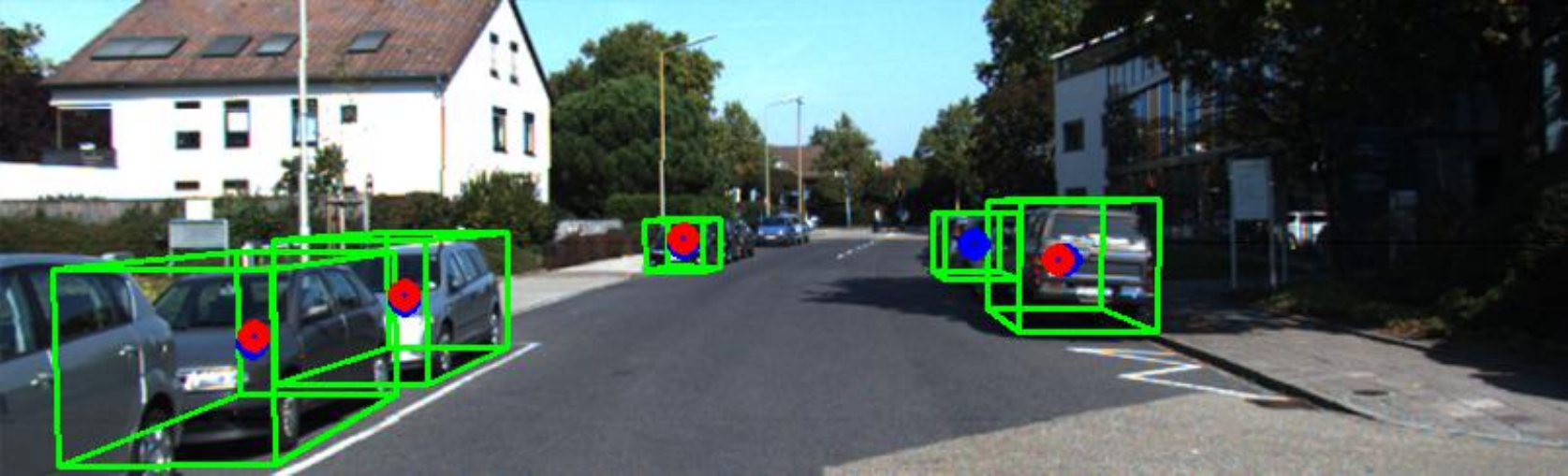} &
         \includegraphics[width=0.3\linewidth]{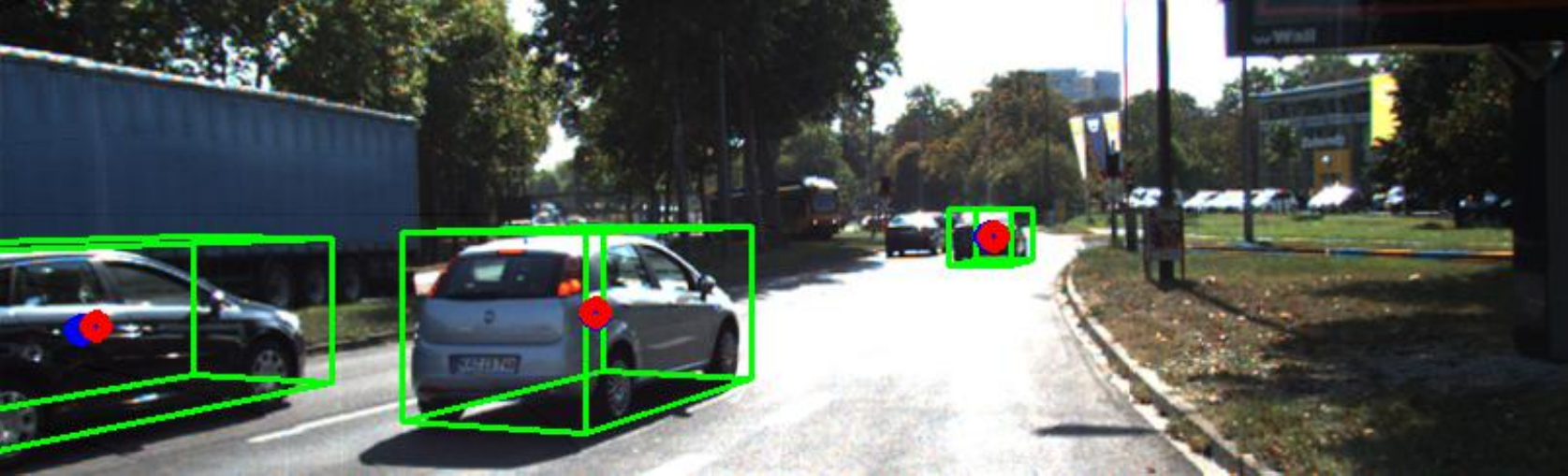} & 
         \includegraphics[width=0.3\linewidth]{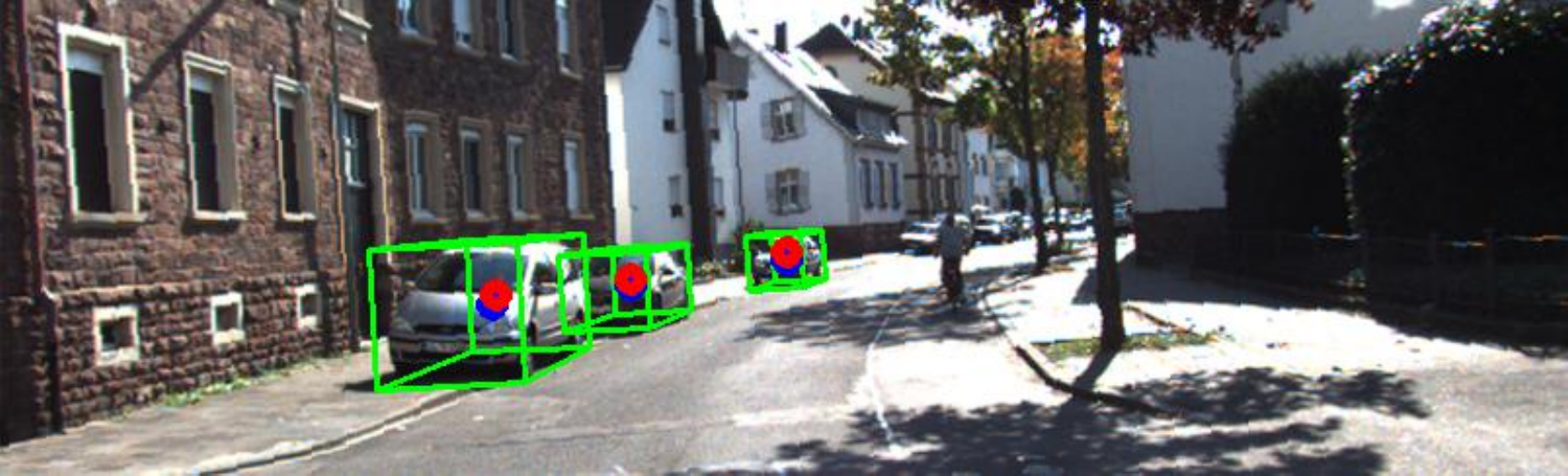} \\ 
         (d) $x$-rotation & (e) $y$-rotation & (f) $z$-rotation \\
    \end{tabular}%
    \vspace{-0.1cm}
    \caption{Qualitative results of MonoDLE (Top) and Ours (Bottom). Projected 3D bounding box (green - prediction) and projected 3D object center (red - GT, blue - prediction) are visualized.}
    \vspace{-2mm}
\label{Qualitative_result_kitti}
\end{figure*}

We generate images as if they are captured from different positions and orientations by applying the basic knowledge of multiple view geometry~\cite{hartley2003multiple} on KITTI datasets~\cite{geiger2012we}.
For image synthesis, we need an image, a per-pixel depth map, and camera parameters.
KITTI provides all of them, but the 3D points are sparse so we need additional dense depth map computation.
We estimate a dense depth map using the state-of-the-art stereo matching network, HITNET~\cite{tankovich2021hitnet}. 
Given depth map $\mathbf{D}\in \mathbb{R}^{H\times W}$, and pre-calibrated camera intrinsic $\mathbf{K}$, we generate a new image $\mathbf{I}_{target}\in \mathbb{R}^{H\times W}$ in \figref{Qualitative_result_kitti}-(b-f)  by back-projecting the reference image $\mathbf{I}_{ref}\in \mathbb{R}^{H\times W}$ in \figref{Qualitative_result_kitti}-(a) into 3D world space, then re-project the 3D points into a new image plane with the relative camera pose $[\mathbf{R}|\mathbf{t}]$ as follows:
\begin{equation}
\begin{gathered}
    \mathbf{I}_{target}(\mathbf{x}) = \mathbf{I}_{ref}(\mathbf{x}'),~\text{where}\\
    \mathbf{x}' = \pi\big(\mathbf{K}[\mathbf{R}|\mathbf{t}]\begin{bmatrix}
\mathbf{X}\\
1
\end{bmatrix}\big),~\mathbf{X} = \big(\mathbf{K}^{-1}\begin{bmatrix}
\mathbf{x}\\
1
\end{bmatrix}\big)\mathbf{D}(\mathbf{x}),\\
\end{gathered}
\end{equation}
where $\mathbf{x}=[x,y]^{\intercal}$ and $\mathbf{X}=[X,Y,Z]^{\intercal}$ are 2D image coordinates and 3D camera coordinates, respectively. The projection function $\pi(\cdot)$ maps 3D points $[a,b,c]$ into 2D pixel coordinates $[a/c,b/c]$.

Since we use the estimated depth from stereo matching, the generated target image $\mathbf{I}_{target}$ contains occlusions and uncertain depth values. We disregard these areas for image warping and fill the holes with a pre-trained image inpainting network~\cite{yu2019free}.
We crop the generated image with $804\times244$ resolution (KITTI original resolution: $1280\times375$) with the same 2D image center to reduce the artifact on the image boundary caused by image warping. 
We generate the target images $\mathbf{I}_{target}$ with the changes in camera translation $\mathbf{t}=[t_x,t_y,t_z]^{\intercal}$ and camera orientation $\mathbf{r}=[r_{pitch}$, $r_{yaw}$, $r_{roll}]$. We consider the KITTI right image to be the generated image translated along the $x$-axis.
We skip generating images with $z$-axis translation because it is widely generalized (\textit{e.g.} the previous and next frames).
We additionally synthesize the images with a combination of both rotation and translation changes.

\subsection{Our 3D Object Detection Method}
\label{sec:ours}
We use the conventional monocular 3D object detection networks, MonoGRNet \cite{qin2019monogrnet}, SMOKE \cite{liu2020smoke}, MonoDLE \cite{ma2021delving}, MonoFLEX \cite{zhang2021objects} trained with the cropped original KITTI datasets.
These methods predict the projected 3D object center $[x_{2d}, y_{2d}]^{\intercal}$, depth $z_{3d}$, yaw angle $r_{yaw}$, and 3D Bounding Box (BB) size $\mathbf{S}=[h,w,l]$, then the object 3D bounding box is computed by \eqnref{eq:objectcenter} and \eqnref{our_method}.
First, the 3D object center $\mathbf{X}_{c}=[X_{c},Y_{c},Z_{c}]$ is computed as follows:
\begin{equation}
\begin{gathered}
    \mathbf{X}_{c} = \begin{bmatrix}
    {X}_{c} \\
    {Y}_{c} \\
    {Z}_{c}
    \end{bmatrix}=z_{3d}\bigg(\mathbf{K}^{-1}\begin{bmatrix}
    {x}_{2d}\\
    {y}_{2d}\\
    1
    \end{bmatrix}\bigg).
\end{gathered}
\label{eq:objectcenter}
\end{equation}
Then, the final output of the object's 8 corners is the eight corner 3D bounding box $\mathbf{B}_{prev}$ computed as follows:
\begin{equation}
\begin{gathered}
    \mathbf{B}_{prev} = \mathbf{R}_{y}(r_{yaw})\begin{bmatrix}
\pm h/2\\
\pm w/2\\
\pm l/2
\end{bmatrix}+\begin{bmatrix}
X_{c}\\
Y_{c}\\
Z_{c}
\end{bmatrix},
\end{gathered}
\label{prev_method}
\end{equation}
where $h, w$, and $l$ are the height, width and length of the 3D bounding box. They represent the 3D heading direction with the yaw axis instead of all orientations (roll, pitch, yaw-axis), which means the camera-to-road relative camera pose is fixed.  
This is a reasonable assumption because the position and angle of the camera system mounted on a vehicle is fixed in the manufacturing process. 
However, this mild assumption causes a drastic performance drop when the model is applied to other camera systems. 
To improve the generalization performance of the models for various camera systems, we design the 3D heading compensation module as shown in \figref{overview}.
Given the relative camera pose $\mathbf{r}_{r\rightarrow t}=[\hat{r}_{roll}, \hat{r}_{pitch}, \hat{r}_{yaw}]$ between a reference camera (training set) and a target camera (test set), we build the final bounding box $\mathbf{B}_{ours}$ as follows:
\begin{equation}
\begin{gathered}
    \mathbf{B}_{ours} = \mathbf{R}_{3d}\begin{bmatrix}
\pm h/2\\
\pm w/2\\
\pm l/2
\end{bmatrix}+\begin{bmatrix}
X_{c}\\
Y_{c}\\
Z_{c}
\end{bmatrix},~\text{where}\\
\mathbf{R}_{3d}=\mathbf{R}_{z}(\hat{r}_{roll})\mathbf{R}_{y}(r_{yaw})\mathbf{R}_{x}(\hat{r}_{pitch}),\\
\end{gathered}
\label{our_method}
\end{equation}
where $\mathbf{R}_{x}(r)$ is the rotation matrix rotating $r$-degree in the $x$-axis.
We use pre-calibrated roll ($\hat{r}_{roll}$) and pitch ($\hat{r}_{pitch}$) angles while the predicted yaw angle ($r_{yaw}$) is utilized from a network, either MonoGRNet, SMOKE, MonoDLE or MonoFLEX.

\begin{table*}[t]
\centering
\begin{footnotesize}
\vspace{0.1cm}
\begin{tabular}{c|c|llllll}
\hline
\multirow{2}{*}{Dataset}                                                  & \multirow{2}{*}{Method} & \multicolumn{3}{c|}{$AP_{3D}(IoU=0.7)$}                            & \multicolumn{3}{c}{$AP_{BEV}$}                    \\ \cline{3-8} 
&                         & E             & M            & \multicolumn{1}{c|}{H} & E             & M             & H             \\ \hline
Original                                                                  & MonoGRNet               & 12.2          & 8.12         & 6.94                   & 21.8          & 17.2          & 13.7          \\ \hline
\multirow{2}{*}{Roll}                                                     & MonoGRNet               & $0.74_{(6\%)}$    & $0.54_{(6\%)}$   & $0.37_{(5\%)}$             & $17.2_{(79\%)}$    & $13.7_{(79\%)}$    & $11.9_{(87\%)}$    \\ 
& \textbf{+Ours}          & $\textbf{10.1}_{(82\%)}$    & $\textbf{6.45}_{(79\%)}$   & $\textbf{4.84}_{(70\%)}$             & $\textbf{18.3}_{(84\%)}$    &$\textbf{16.1}_{(94\%)}$    & $\textbf{13.5}_{(99\%)}$    \\ \hline
\multirow{2}{*}{Pitch}                                                    & MonoGRNet               & $1.29_{(11\%)}$    & $0.81_{(10\%)}$   & $0.57_{(8\%)}$             & $22.9_{(104\%)}$    & $18.0_{(104\%)}$    & $14.1_{(103\%)}$    \\ 
& \textbf{+Ours}          & $\textbf{8.15}_{(67\%)}$    & $\textbf{5.89}_{(73\%)}$   & $\textbf{5.12}_{(74\%)}$             & $\textbf{24.1}_{(110\%)}$    & $\textbf{19.7}_{(114\%)}$    & $\textbf{15.9}_{(116\%)}$    \\ \hline
\multirow{2}{*}{Yaw}                                                      & MonoGRNet               & $8.39_{(69\%)}$    & $5.31_{(65\%)}$   & $4.31_{(62\%)}$             & $21.7_{(99\%)}$    & $16.8_{(97\%)}$    & $13.2_{(96\%)}$    \\ 
& \textbf{+Ours}          & $\textbf{9.54}_{(78\%)}$    & $\textbf{6.37}_{(78\%)}$   & $\textbf{5.5}_{(79\%)}$             & $\textbf{22.2}_{(102\%)}$    & $\textbf{16.9}_{(98\%)}$    & $\textbf{13.1}_{(97\%)}$    \\ \hline
\multirow{2}{*}{TransX}                                                   & MonoGRNet               & $10.7_{(88\%)}$          & $7.89_{(97\%)}$          & $6.17_{(89\%)}$                    & $19.0_{(87\%)}$            & $15.5_{(90\%)}$         & $13.1_{(96\%)}$         \\
& \textbf{+Ours}          & $\textbf{10.7}_{(88\%)}$ & $\textbf{7.89}_{(97\%)}$ & $\textbf{6.17}_{(89\%)}$           & $\textbf{19.0}_{(87\%)}$   & $\textbf{15.5}_{(90\%)}$ & $\textbf{13.1}_{(96\%)}$ \\ \hline
\multirow{2}{*}{TransY}                                                   & MonoGRNet               & $11.1_{(91\%)}$           & $7.71_{(95\%)}$          & $6.37_{(92\%)}$                     & $18.1_{(83\%)}$           & $14.2_{(83\%)}$           & $12.8_{(93\%)}$           \\
& \textbf{+Ours}          & $\textbf{11.1}_{(91\%)}$ & $\textbf{7.71}_{(95\%)}$ & $\textbf{6.37}_{(92\%)}$           & $\textbf{18.1}_{(83\%)}$   & $\textbf{14.2}_{(83\%)}$ & $\textbf{12.8}_{(93\%)}$\\ \hline
\multirow{2}{*}{\begin{tabular}[c]{@{}c@{}}TransY\\ + Pitch\end{tabular}} & MonoGRNet               & $1.09_{(9\%)}$    & $0.76_{(9\%)}$   & $0.53_{(8\%)}$             & $17.0_{(78\%)}$    & \textbf[$14.7_{(85\%)}$    & $13.1_{(96\%)}$    \\ 
& \textbf{+Ours}          & $\textbf{8.12}_{(67\%)}$    & $\textbf{6.14}_{(79\%)}$   & $\textbf{5.45}_{(78\%)}$             & $\textbf{17.8}_{(81\%)}$    & $\textbf{16.0}_{(93\%)}$    & $\textbf{14.7}_{(107\%)}$    \\ \hline\hline
Original                                                                  & SMOKE               & 16.58          & 9.56         & 9.12                   &  18.5         & 14.3         & 13.9          \\ \hline
\multirow{2}{*}{Roll}                                                     & SMOKE               & $0.88_{(5\%)}$    & $0.61_{(6\%)}$   & $0.50_{(5\%)}$             & $17.3_{(93\%)}$    & $13.9_{(97\%)}$    & $12.6_{(91\%)}$    \\ 
& \textbf{+Ours}          & $\textbf{13.54}_{(81\%)}$    & $\textbf{7.40}_{(77\%)}$   & $\textbf{6.11}_{(67\%)}$           & $\textbf{18.3}_{(99\%)}$    & $\textbf{17.1}_{(119\%)}$    & $\textbf{15.6}_{(108\%)}$    \\ \hline
\multirow{2}{*}{Pitch}                                                    & SMOKE               & $1.54_{(9\%)}$    & $1.01_{(10\%)}$   & $0.76_{(8\%)}$             & $19.2_{(104\%)}$    & $15.4_{(108\%)}$    & $14.0_{(100\%)}$    \\ 
& \textbf{+Ours}          & $\textbf{9.89}_{(60\%)}$    & $\textbf{6.56}_{(69\%)}$   & $\textbf{6.42}_{(70\%)}$             & $\textbf{20.9}_{(113\%)}$    & $\textbf{16.8}_{(118\%)}$    & $\textbf{15.8}_{(114\%)}$    \\ \hline
\multirow{2}{*}{Yaw}                                                      & SMOKE               & $11.7_{(70\%)}$    & $6.34_{(66\%)}$   & $5.26_{(58\%)}$             & $17.2_{(93\%)}$    & $14.1_{(98\%)}$    & $13.4_{(96\%)}$    \\ 
& \textbf{+Ours}          & $\textbf{13.8}_{(83\%)}$    & $\textbf{7.54}_{(79\%)}$   & $\textbf{6.94}_{(76\%)}$             & $\textbf{17.8}_{(96\%)}$    & $\textbf{14.1}_{(98\%)}$    & $\textbf{13.6}_{(98\%)}$    \\ \hline
\multirow{2}{*}{TransX}                                                   & SMOKE                   & $14.1_{(85\%)}$          & $9.56_{(89\%)}$          & $9.12_{(87\%)}$                    & $18.5_{(98\%)}$            & $14.3_{(102\%)}$         & $13.9_{(94\%)}$          \\
& \textbf{+Ours}          & $\textbf{14.1}_{(85\%)}$ & $\textbf{9.56}_{(89\%)}$ & $\textbf{9.12}_{(87\%)}$           & $\textbf{18.5}_{(98\%)}$   & $\textbf{14.3}_{(102\%)}$ & $\textbf{13.9}_{(94\%)}$ \\ \hline
\multirow{2}{*}{TransY}                                                   & SMOKE                   & $14.8_{(89\%)}$          & $9.11_{(95\%)}$          & $8.56_{(87\%)}$                    & $19.0_{(102\%)}$            & $15.4_{(108\%)}$         & $12.8_{(92\%)}$          \\
& \textbf{+Ours}          & $\textbf{14.8}_{(89\%)}$ & $\textbf{9.11}_{(95\%)}$ & $\textbf{8.56}_{(87\%)}$           & $\textbf{19.0}_{(102\%)}$   & $\textbf{15.4}_{(108\%)}$ & $\textbf{12.8}_{(92\%)}$ \\ \hline
\multirow{2}{*}{\begin{tabular}[c]{@{}c@{}}TransY\\ + Pitch\end{tabular}} & SMOKE               & $1.33_{(8\%)}$    & $0.89_{(9\%)}$   & $0.61_{(7\%)}$             & $17.0_{(92\%)}$    & $13.9_{(97\%)}$    & $12.0_{(86\%)}$    \\ 
& \textbf{+Ours}          & $\textbf{10.4}_{(63\%)}$    & $\textbf{6.44}_{(67\%}$)   & $\textbf{5.64}_{(62\%)}$             & $\textbf{18.1}_{(98\%)}$    & $\textbf{15.0}_{(105\%)}$    & $\textbf{13.7}_{(99\%)}$    \\ \hline\hline
Original                                                                  & MonoDLE               & 13.6          & 11.3         & 9.69                   & 19.7          &  16.1         & 14.7          \\ \hline
\multirow{2}{*}{Roll}                                                     & MonoDLE               & $1.29_{(9\%)}$    & $0.99_{(9\%)}$   & $0.92_{(9\%)}$             & $17.3_{(88\%)}$    & $14.5_{(90\%)}$    & $12.7_{(86\%)}$    \\ 
& \textbf{+Ours}          & $\textbf{11.8}_{(87\%)}$    & $\textbf{9.10}_{(80\%)}$   & $\textbf{7.98}_{(82\%)}$           & $\textbf{18.3}_{(93\%)}$    & $\textbf{16.7}_{(103\%)}$    & $\textbf{15.0}_{(103\%)}$    \\ \hline
\multirow{2}{*}{Pitch}                                                    & MonoDLE               & $1.96_{(14\%)}$    & $1.76_{(16\%)}$   & $1.45_{(15\%)}$             & $20.0_{(102\%)}$    & $14_{(88\%)}$    & $12.8_{(87\%)}$    \\ 
& \textbf{+Ours}          & $\textbf{9.91}_{(73\%)}$    & $\textbf{9.12}_{(81\%)}$   & $\textbf{7.82}_{(81\%)}$             & $\textbf{21.9}_{(111\%)}$    & $\textbf{16.5}_{(102\%)}$    & $\textbf{16.0}_{(109\%)}$    \\ \hline
\multirow{2}{*}{Yaw}                                                      & MonoDLE               & $9.42_{(69\%)}$    & $6.91_{(60\%)}$   & $5.72_{(59\%)}$             & $20.2_{(102\%)}$    & $17.8_{(110\%)}$    & $16.9_{(115\%)}$    \\ 
& \textbf{+Ours}          & $\textbf{10.7}_{(79\%)}$    & $\textbf{9.44}_{(83\%)}$   & $\textbf{7.87}_{(81\%)}$             & $\textbf{21.6}_{(110\%)}$    & $\textbf{18.9}_{(117\%)}$    & $\textbf{17.4}_{(119\%)}$    \\ \hline
\multirow{2}{*}{TransX}                                                   & MonoDLE                 & $13.6_{(96\%)}$          & $10.1_{(89\%)}$          & $8.89_{(92\%)}$                    & $19.4_{(98\%)}$            & $15.6_{(97\%)}$         & $14.5_{(99\%)}$          \\
& \textbf{+Ours}          & $\textbf{13.6}_{(96\%)}$ & $\textbf{10.1}_{(89\%)}$ & $\textbf{8.89}_{(92\%)}$           & $\textbf{19.4}_{(98\%)}$   & $\textbf{15.6}_{(97\%)}$ & $\textbf{14.5}_{(99\%)}$  \\ \hline
\multirow{2}{*}{TransY}                                                   & MonoDLE                 & $13.8_{(101\%)}$          & $10.8_{(96\%)}$          & $8.56_{(88\%)}$                    & $19.2_{(97\%)}$            & $15.4_{(95\%)}$         & $14.4_{(98\%)}$          \\
& \textbf{+Ours}          & $\textbf{13.8}_{(101\%)}$ & $\textbf{10.8}_{(96\%)}$ & $\textbf{8.56}_{(88\%)}$           & $\textbf{19.2}_{(97\%)}$   & $\textbf{15.4}_{(95\%)}$ & $\textbf{14.4}_{(98\%)}$  \\ \hline
\multirow{2}{*}{\begin{tabular}[c]{@{}c@{}}TransY\\ + Pitch\end{tabular}} & MonoDLE               & $1.81_{(13\%)}$    & $1.44_{(13\%)}$   & $1.29_{(13\%)}$             & $17.5_{(89\%)}$    & $15.0_{(93\%)}$    & $12.8_{(88\%)}$    \\ 
& \textbf{+Ours}          & $\textbf{9.89}_{(73\%)}$    & $\textbf{9.13}_{(80\%}$)   & $\textbf{7.21}_{(61\%)}$             & $\textbf{18.5}_{(94\%)}$    & $\textbf{16.4}_{(102\%)}$    & $\textbf{14.9}_{(101\%)}$    \\ \hline\hline
Original                                                                  & MonoFLEX               & 14.2          & 9.94         & 7.09                  & 19.67          & 16.11          & 14.67          \\ \hline
\multirow{2}{*}{Roll}                                                     & MonoFLEX               & $1.13_{(8\%)}$    & $0.87_{(9\%)}$   & $7.09_{(10\%)}$             & $16.8_{(83\%)}$    & $14.1_{(88\%)}$    & $12.0_{(83\%)}$    \\ 
& \textbf{+Ours}          & $\textbf{11.0}_{(77\%)}$    & $\textbf{7.12}_{(71\%)}$   & $\textbf{5.45}_{(77\%)}$           & $\textbf{18.7}_{(93\%)}$    & $\textbf{16.9}_{(106\%)}$    & $\textbf{14.3}_{(97\%)}$    \\ \hline
\multirow{2}{*}{Pitch}                                                    & MonoFLEX               & $2.12_{(15\%)}$    & $1.34_{(13\%)}$   & $0.99_{(14\%)}$             & $21.0_{(104\%)}$    & $13.9_{(87\%)}$    & $12.9_{(90\%)}$    \\ 
& \textbf{+Ours}          & $\textbf{9.89}_{(60\%)}$    & $\textbf{6.56}_{(69\%)}$   & $\textbf{6.42}_{(70\%)}$             & $\textbf{20.9}_{(113\%)}$    & $\textbf{16.8}_{(118\%)}$    & $\textbf{15.8}_{(114\%)}$    \\ \hline
\multirow{2}{*}{Yaw}                                                      & MonoFLEX               & $9.8_{(69\%)}$    & $6.44_{(65\%)}$   & $4.23_{60\%}$             & $20.2_{(103\%)}$    & $15.8_{(98\%)}$    & $13.8_{(94\%)}$    \\ 
& \textbf{+Ours}          & $\textbf{12.1}_{(85\%)}$    & $\textbf{8.12}_{(82\%)}$   & $\textbf{5.41}_{(76\%)}$             & $\textbf{20.8}_{(106\%)}$    & $\textbf{16.7}_{(104\%)}$    & $\textbf{14.8}_{(101\%)}$    \\ \hline
\multirow{2}{*}{TransX}                                                   & MonoFLEX                & $14.1_{(99\%)}$          & $9.51_{(96\%)}$          & $7.49_{(106\%)}$                    & $19.3_{(98\%)}$            & $15.7_{(97\%)}$         & $14.4_{(98\%)}$          \\
& \textbf{+Ours}          & $\textbf{14.1}_{(99\%)}$ & $\textbf{9.51}_{(96\%)}$ & $\textbf{7.49}_{(106\%)}$           & $\textbf{19.3}_{(98\%)}$   & $\textbf{15.7}_{(97\%)}$ & $\textbf{14.4}_{(98\%)}$ \\ \hline
\multirow{2}{*}{TransY}                                                   & MonoFLEX                & $14.8_{(104\%)}$          & $9.11_{(92\%)}$          & $7.56_{(107\%)}$                    & $19.5_{(99\%)}$            & $15.5_{(99\%)}$         & $14.2_{(97\%)}$          \\
& \textbf{+Ours}          & $\textbf{14.8}_{(104\%)}$ & $\textbf{9.11}_{(92\%)}$ & $\textbf{7.56}_{(107\%)}$           & $\textbf{19.5}_{(99\%)}$   & $\textbf{15.9}_{(99\%)}$ & $\textbf{14.2}_{(97\%)}$ \\ \hline
\multirow{2}{*}{\begin{tabular}[c]{@{}c@{}}TransY\\ + Pitch\end{tabular}} & MonoFLEX               & $1.94_{(14\%)}$    & $1.27_{(12\%)}$   & $1.09_{(15\%)}$             & $18.8_{(93\%)}$    & $12.1_{(76\%)}$    & $11.1_{(76\%)}$    \\ 
& \textbf{$+Ours$}          & $\textbf{9.57}_{(67\%)}$    & $\textbf{8.54}_{(86\%)}$   & $\textbf{5.89}_{(83\%)}$             & $\textbf{19.7}_{(98\%)}$    & $\textbf{13.8}_{(86\%)}$    & $\textbf{12.7}_{(88\%)}$    \\ \hline\hline
\end{tabular}
\caption{\textbf{Part of quantitative 3D detection results.} An example is improvement when degree is 3. Subscript parentheses indicate the percentage of performance compared to the original dataset performance. E, M, H means easy, moderate and hard, respectively.}
\label{Part_3D_results}
\end{footnotesize}
\end{table*}

\section{Experiments}

\begin{figure*}[t]
    \centering
    \footnotesize
    \resizebox{\textwidth}{!}{
    \begin{tabular}{c@{\hspace{0.1mm}}c@{\hspace{0.1mm}}c@{\hspace{0.1mm}}c@{\hspace{0.1mm}}}
         \includegraphics[width=0.23\linewidth]{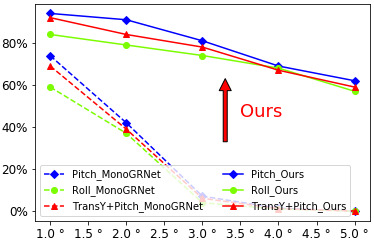}&
         \includegraphics[width=0.23\linewidth]{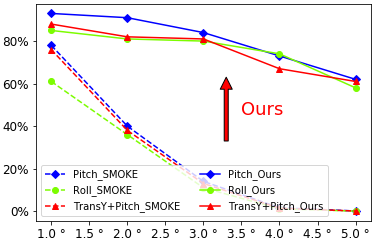}&
         \includegraphics[width=0.23\linewidth]{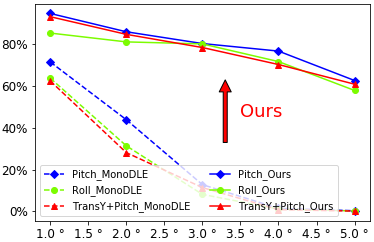}&
         \includegraphics[width=0.23\linewidth]{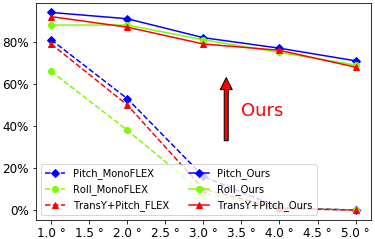} \\
         (a) MonoGRNet & (b) SMOKE & (c) MonoDLE & (d) MonoFLEX
    \end{tabular}
    }
    \vspace{-0.1cm}
    \caption{3D object detection results. The $x$-axis is the rotation angle and the $y$-axis is the performance drop ratio ($\%$) compared to the $AP_{3D}$ (KITTI moderate, IoU=0.7) result of the original image.}
    \vspace{-2mm}
\label{3D_results_graph}
\end{figure*}

\subsection{Comparison to state-of-the-art methods}
\label{sec:state}
We compare our method to state-of-the-art methods, MonoGRNet \cite{qin2019monogrnet}, SMOKE \cite{liu2020smoke}, MonoDLE \cite{ma2021delving}, and MonoFLEX \cite{zhang2021objects}.
All the networks are trained using the cropped original KITTI left images $\mathbf{I}_{ref}$ as shown in \figref{Qualitative_result_kitti}-(a).
The object 3D bounding boxes in \figref{Qualitative_result_kitti} are predicted by passing the various rotated images and translated images $\mathbf{I}_{target}$ generated by following \secref{sec:datageneration} through the trained networks.
We investigate the performance drop of the baseline networks given the images captured at different orientations and positions.
The qualitative results are shown in \figref{Qualitative_result_kitti}-(b-f) and the quantitative results are reported in \figref{3D_results_graph} and \tabref{Part_3D_results}.
We report both $AP_{3D}~(IoU=0.7)$ and $AP_{BEV}$ in \tabref{Part_3D_results}. 
We observe that the $AP_{BEV}$ of conventional methods dropped slightly because it measures the average precision in a 2D Bird-Eye-View projected space. 
All results analysis is conducted based on $AP_{3D}$.
Overall, the translation changes rarely degrade performance in all of the state-of-the-art methods. However, rotation changes, especially in roll and pitch axes, sharply lower the average precision of all competing methods, while the proposed method retains its performance.

\begin{figure*}[t]
    \centering
    \footnotesize
    \begin{tabular}{c@{\hspace{1mm}}c@{\hspace{1mm}}c@{\hspace{1mm}}c@{\hspace{1mm}}}
         \includegraphics[width=0.2\linewidth]{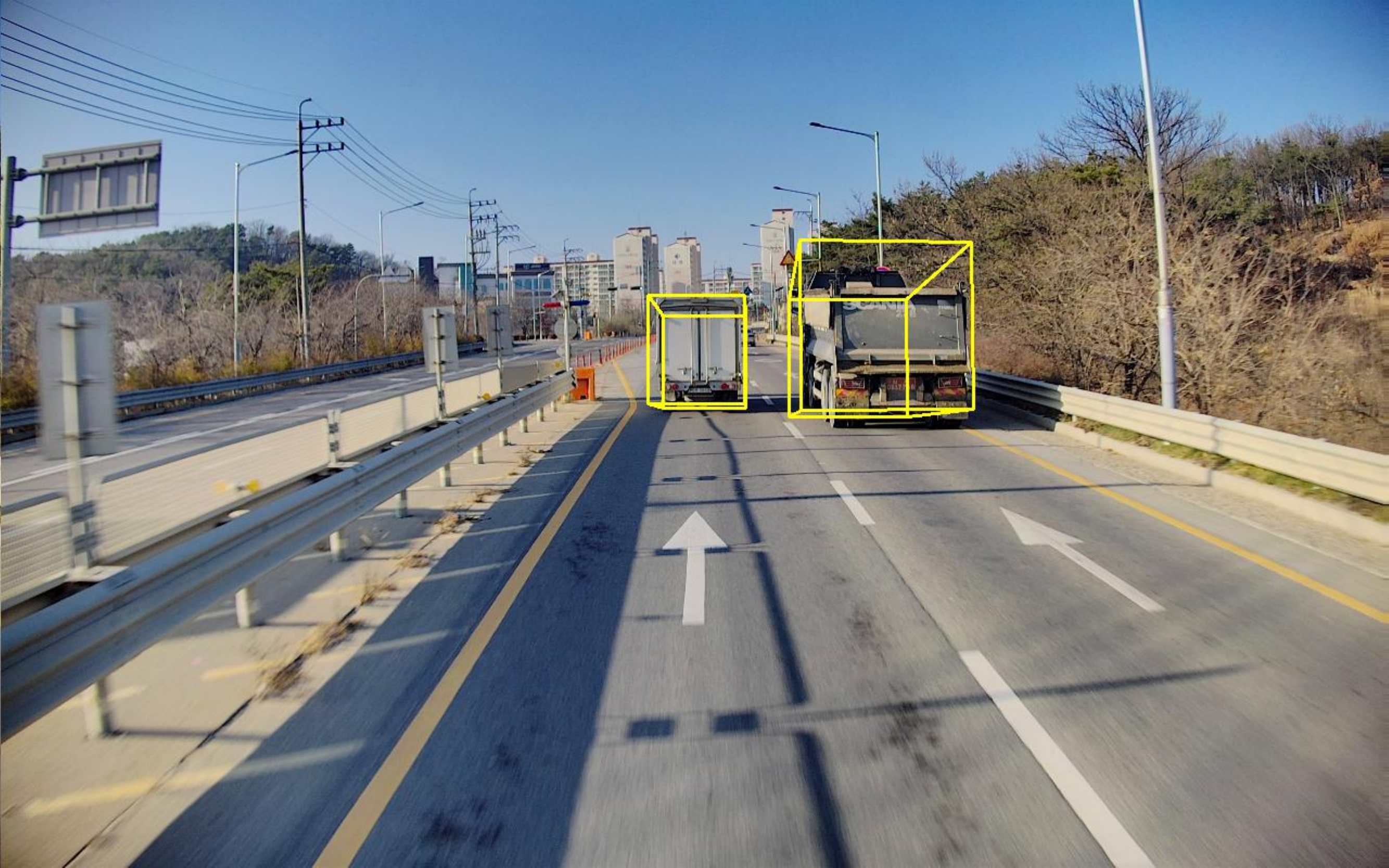}&
         \includegraphics[width=0.2\linewidth]{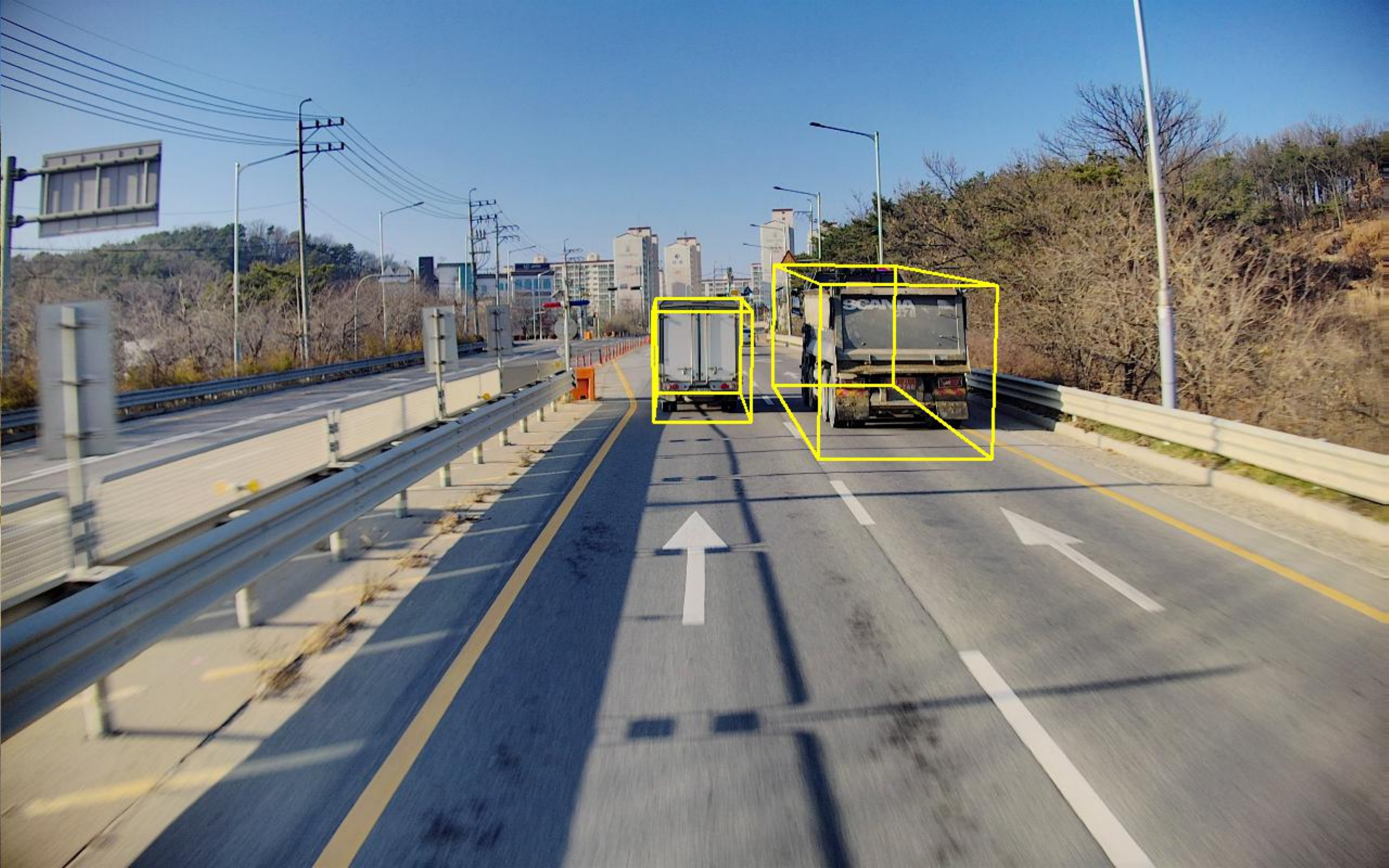}&
         \includegraphics[width=0.2\linewidth]{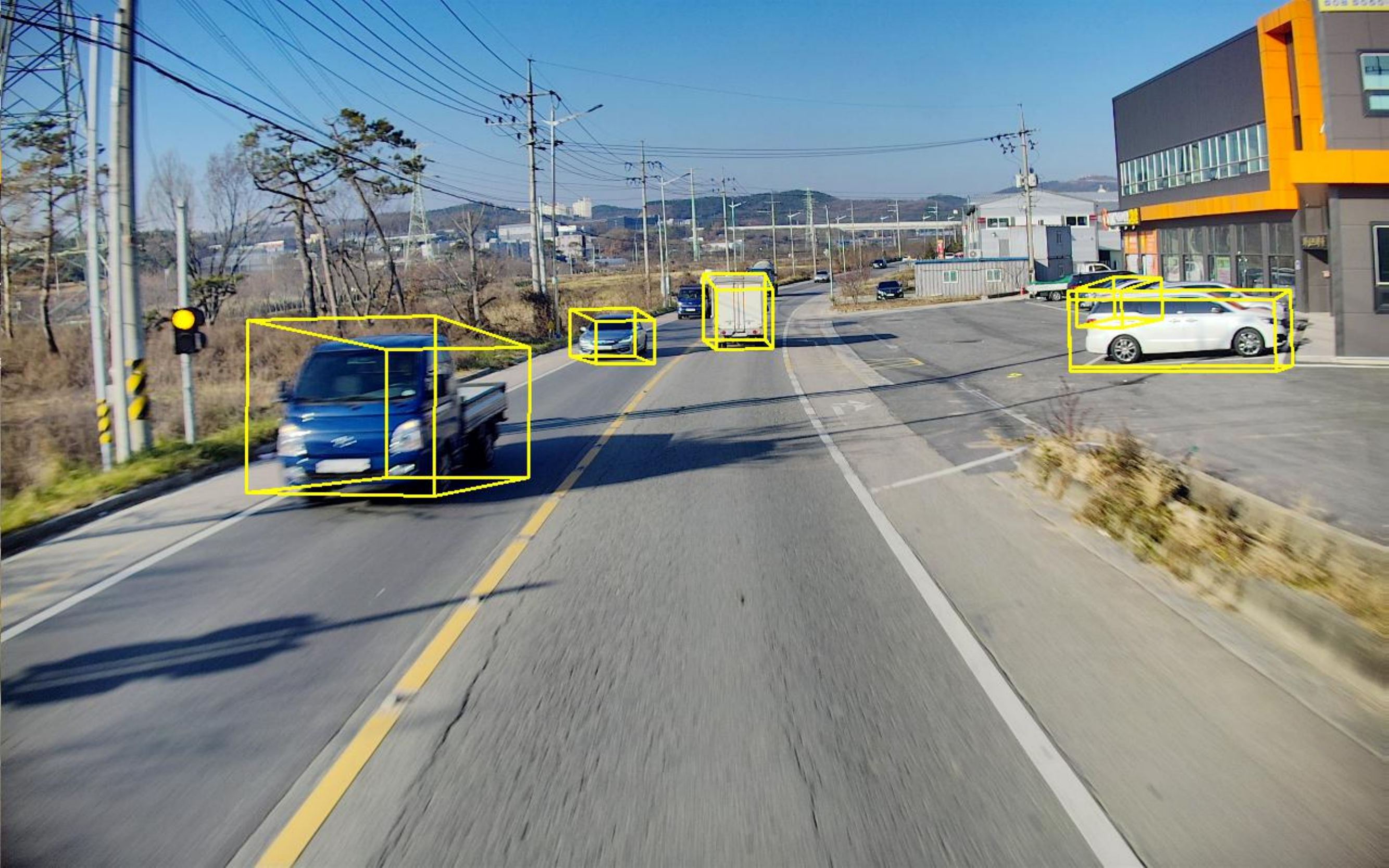}&
         \includegraphics[width=0.2\linewidth]{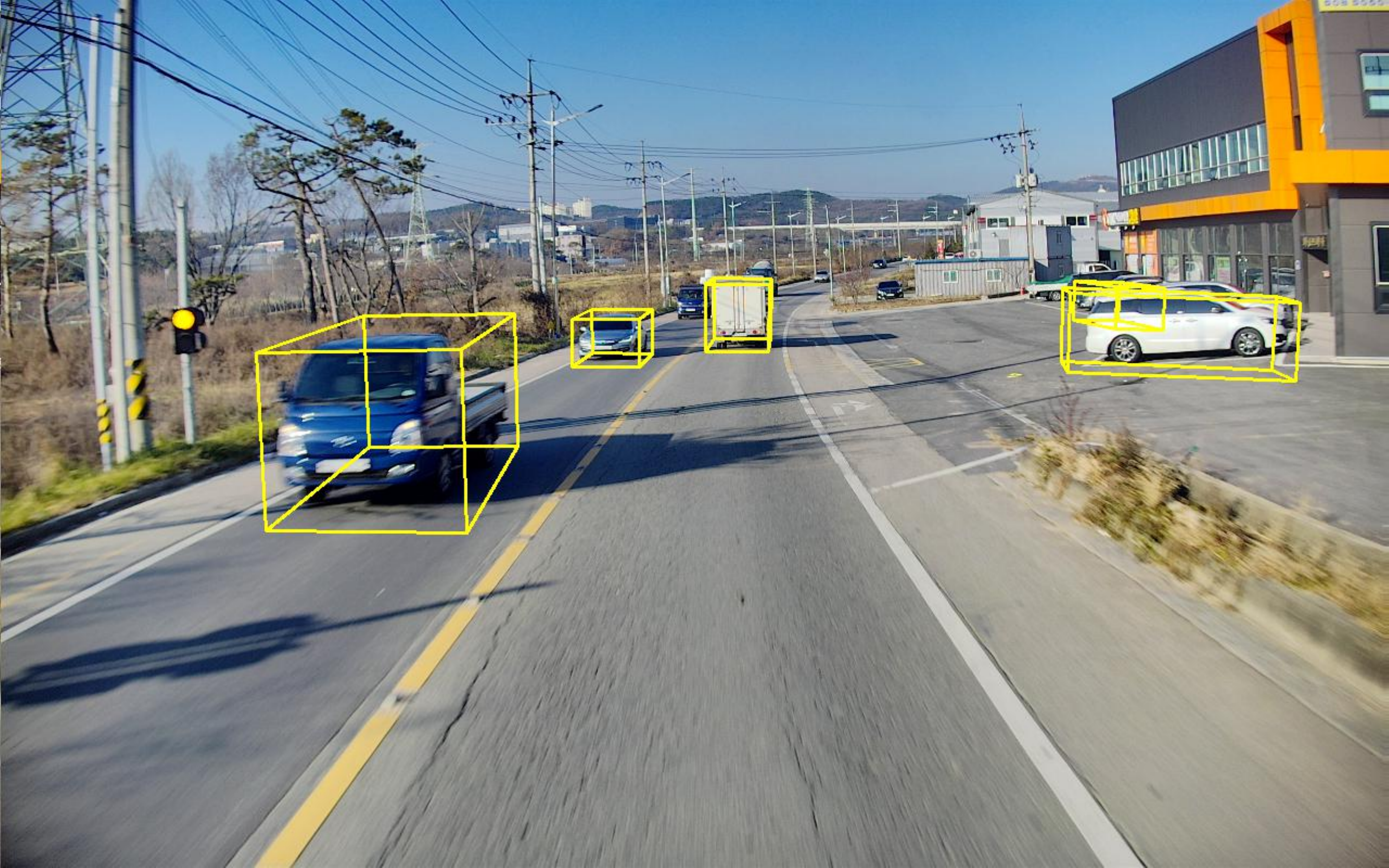} \\
         \includegraphics[width=0.2\linewidth]{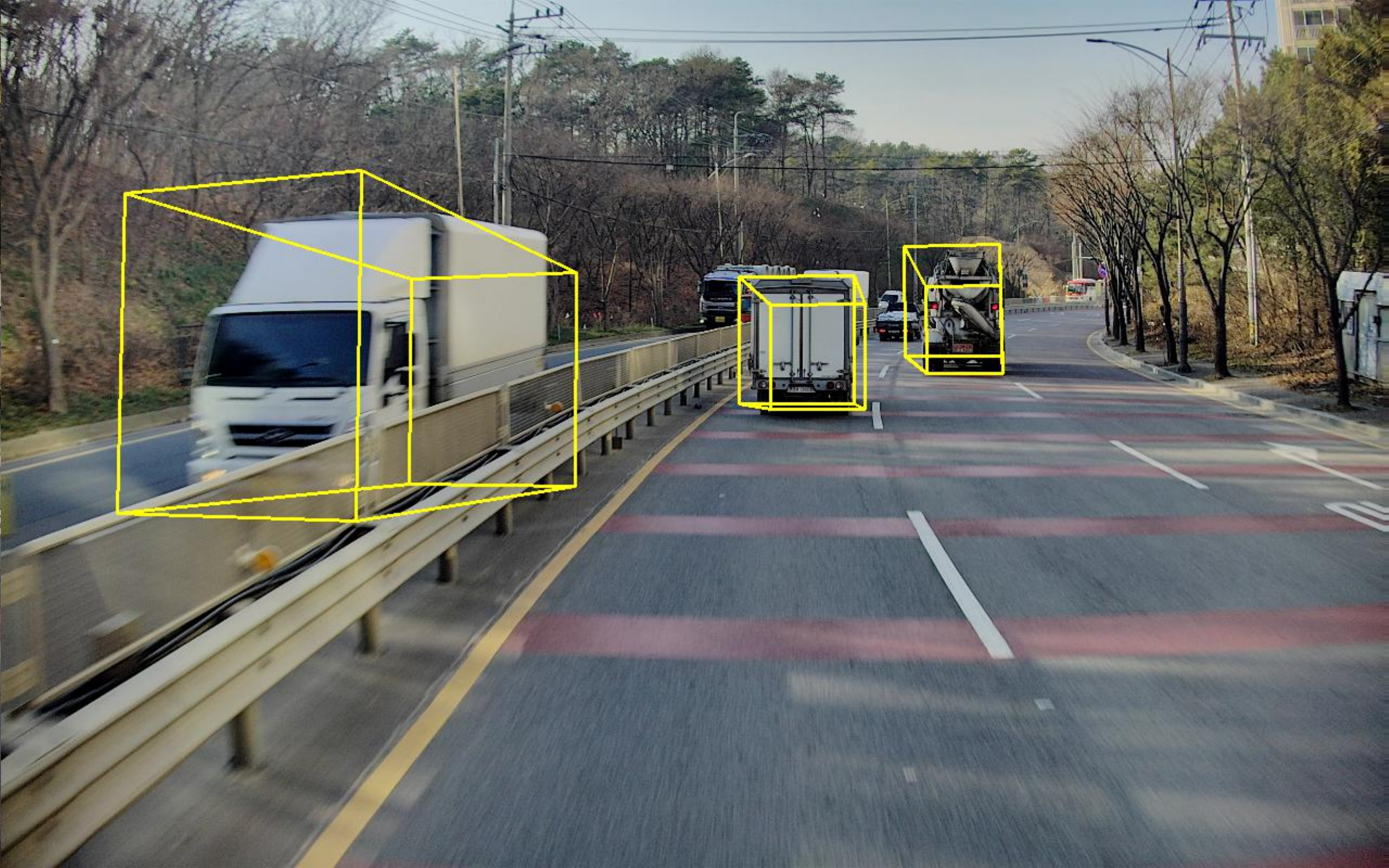}&
         \includegraphics[width=0.2\linewidth]{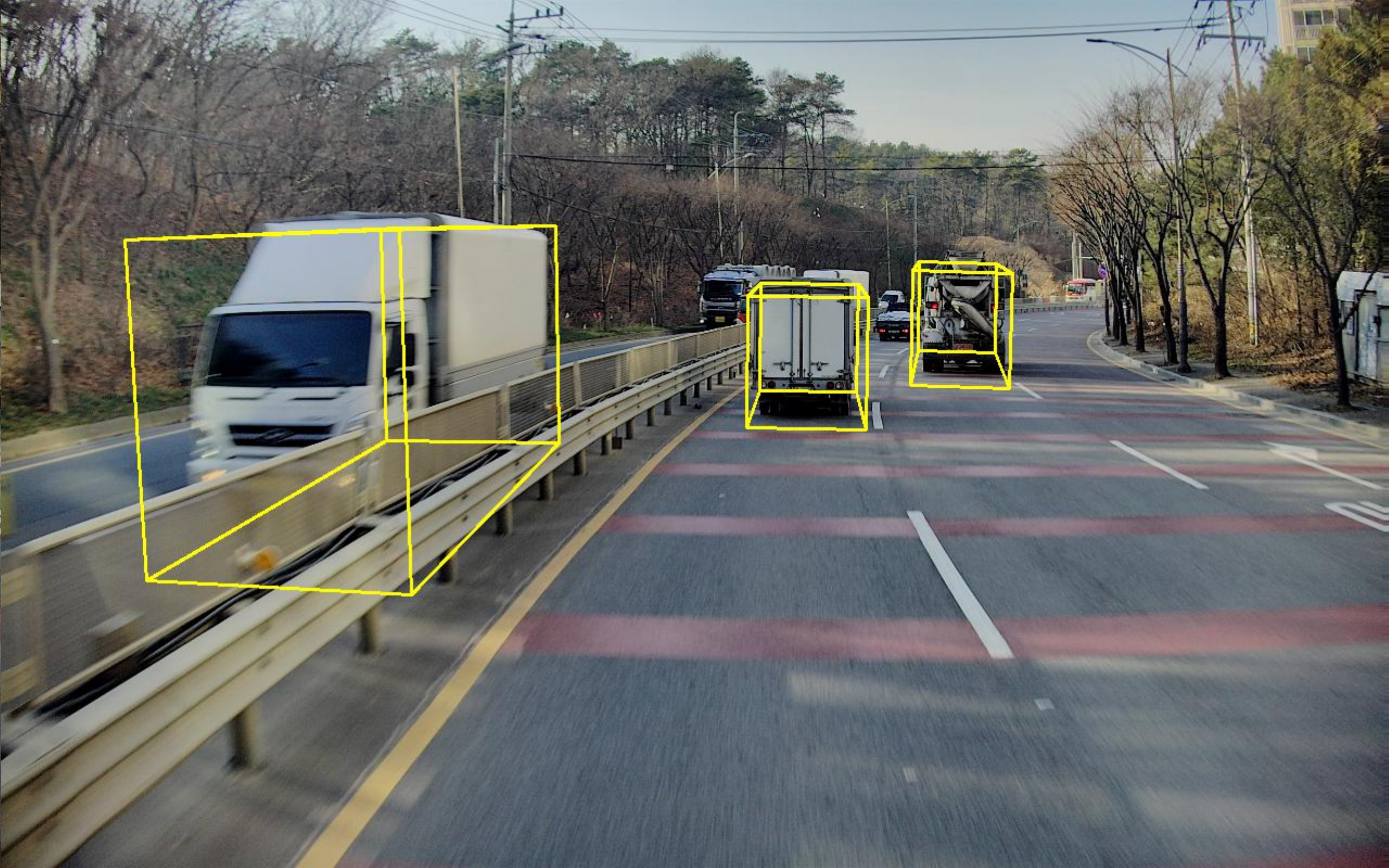}&
         \includegraphics[width=0.2\linewidth]{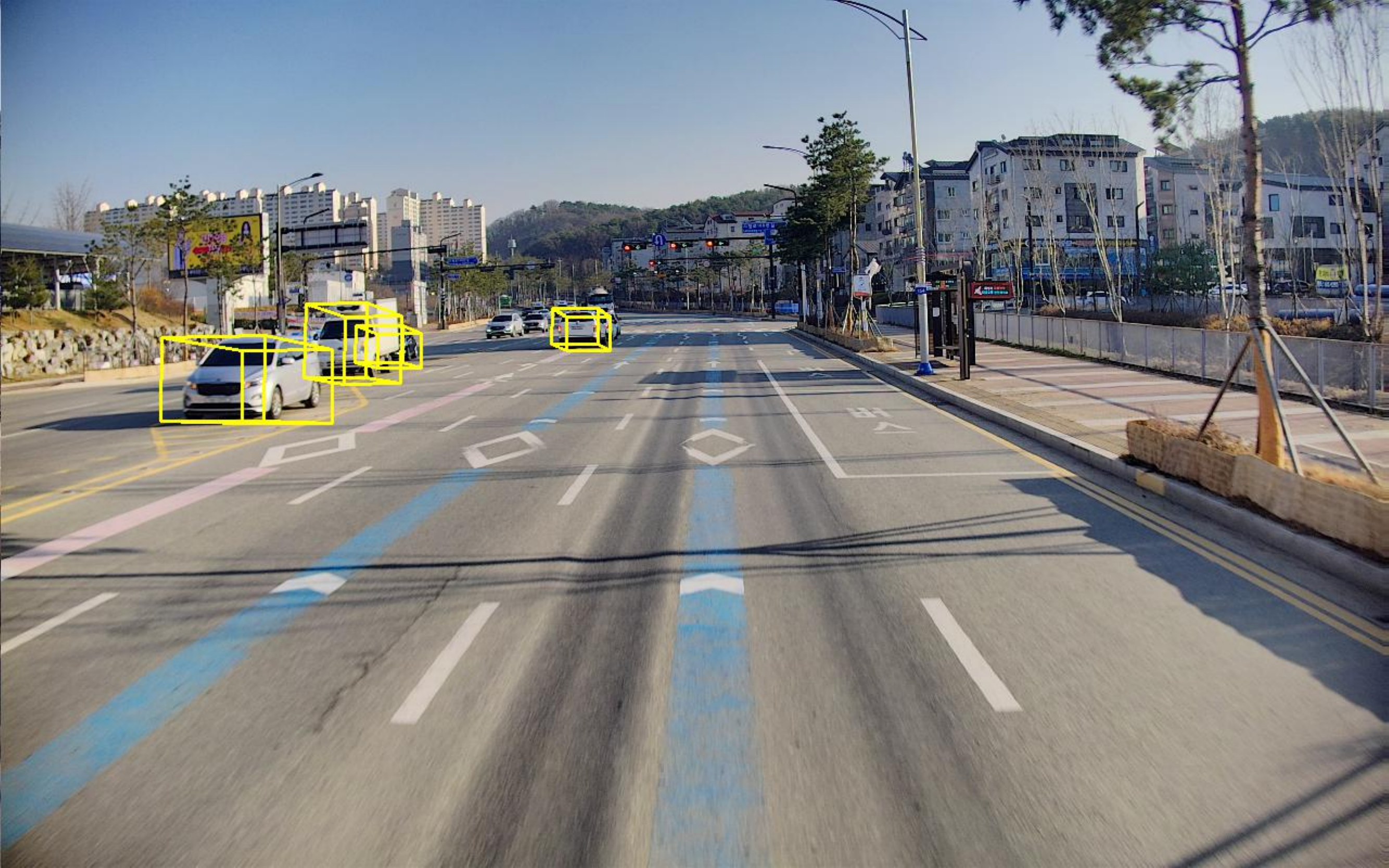}&
         \includegraphics[width=0.2\linewidth]{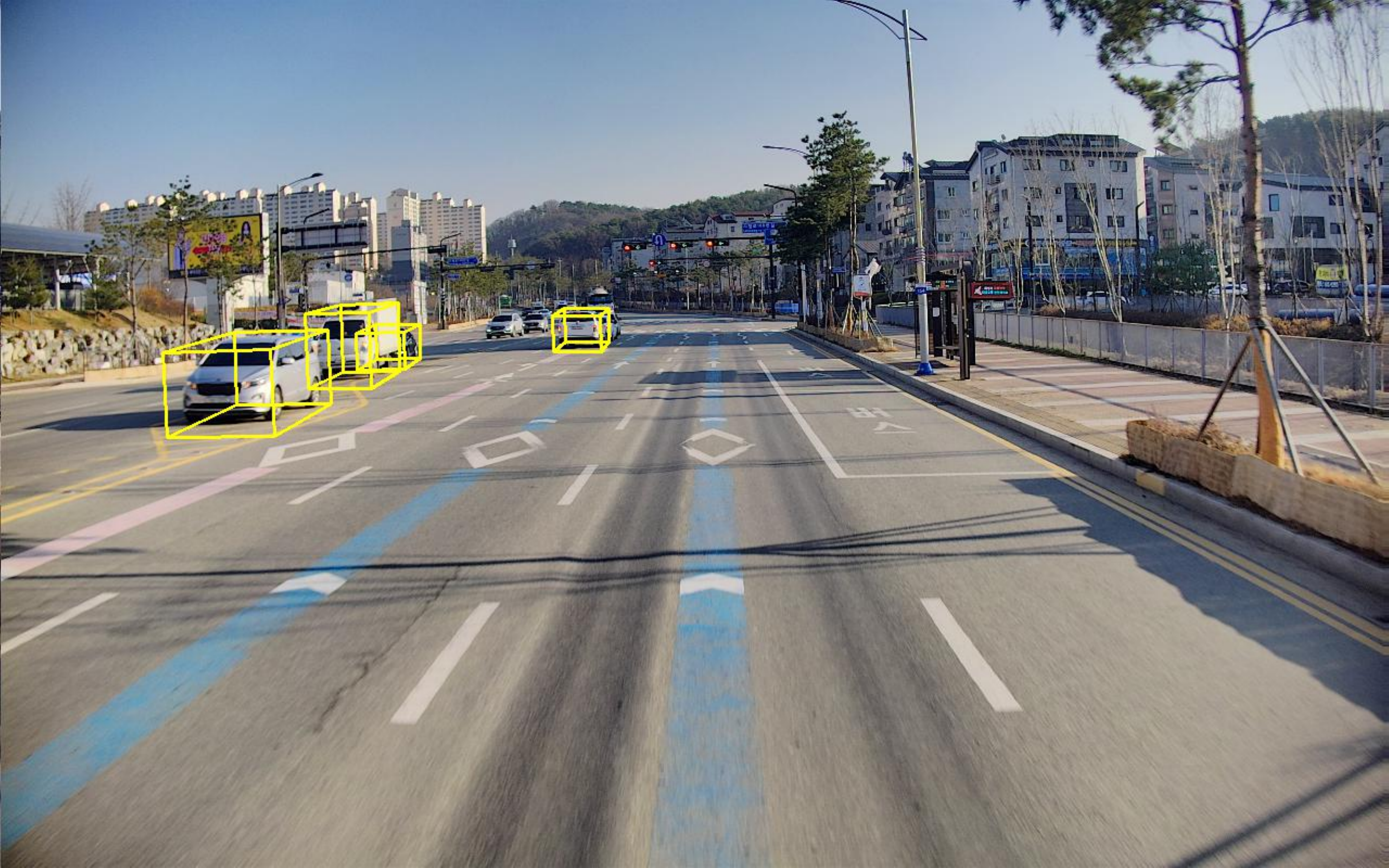} \\
         \includegraphics[width=0.2\linewidth]{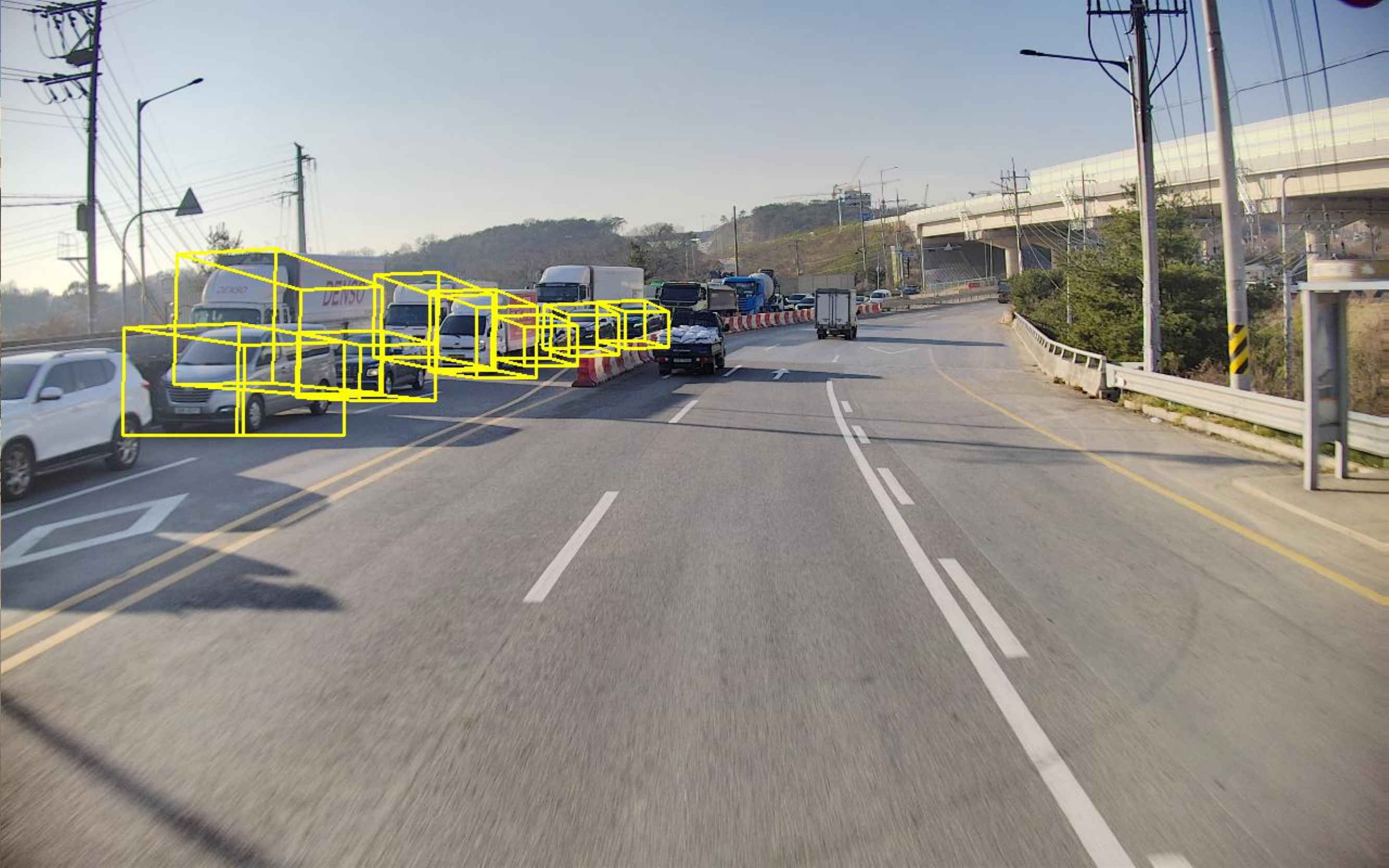}&
         \includegraphics[width=0.2\linewidth]{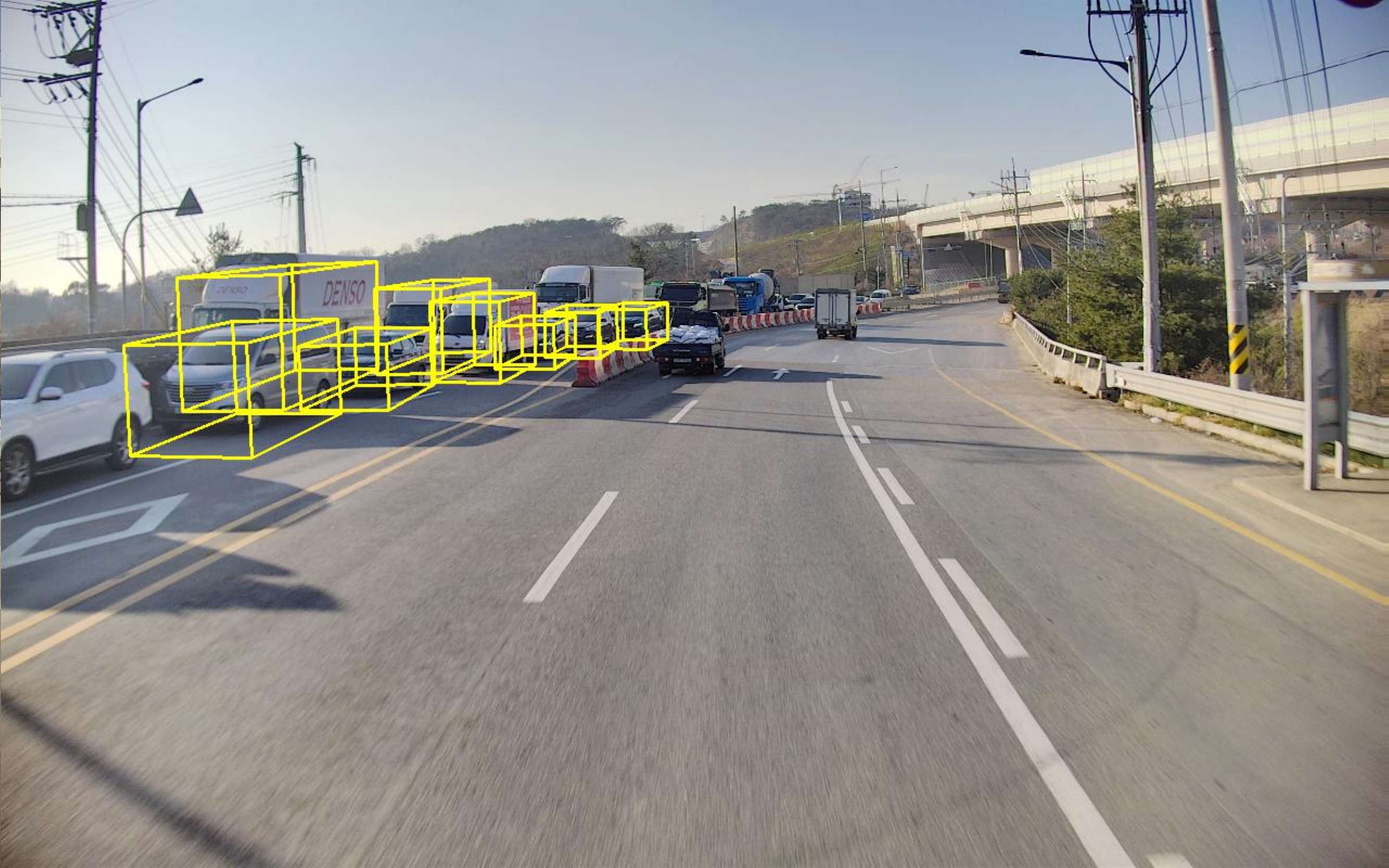}&
         \includegraphics[width=0.2\linewidth]{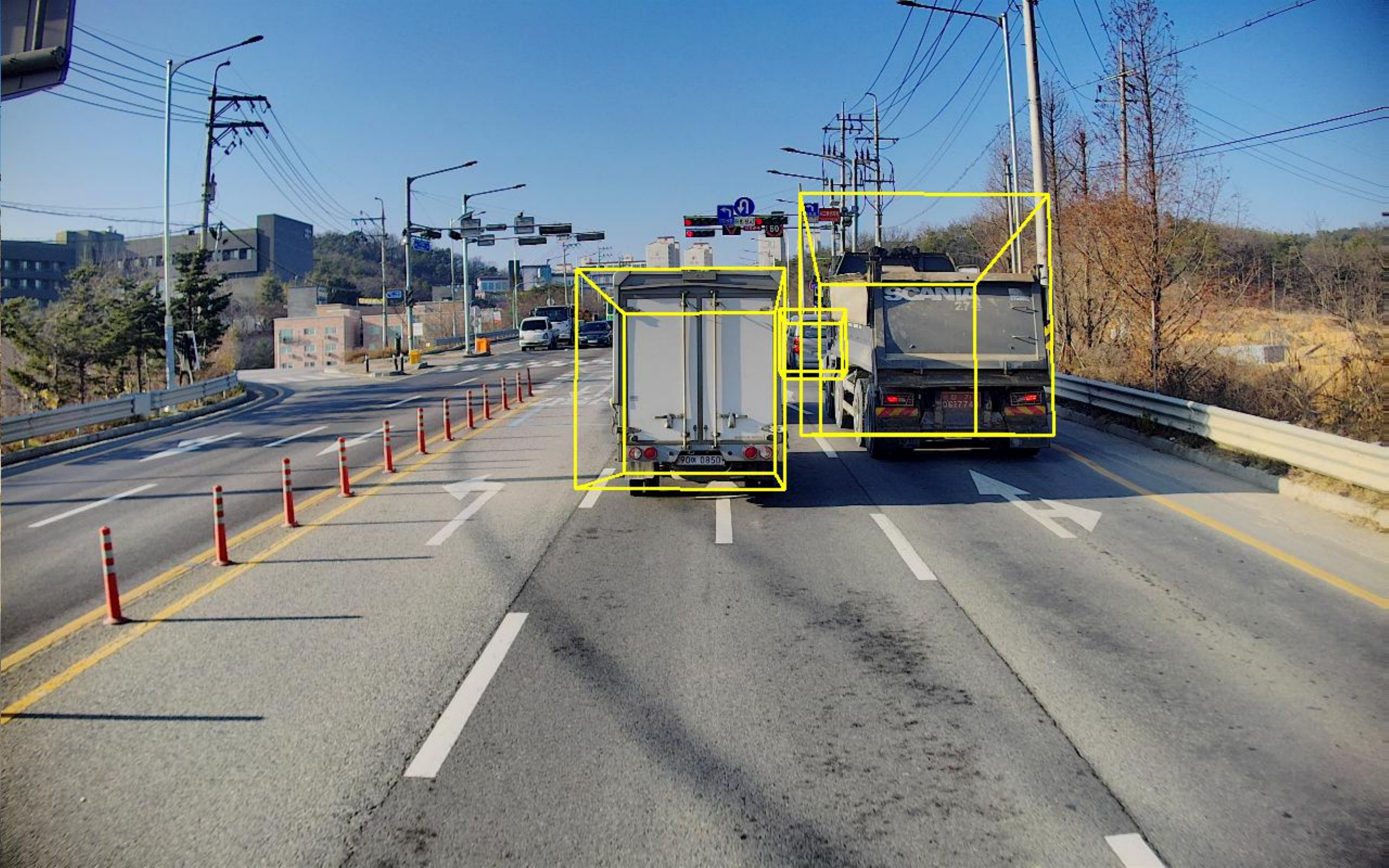}&
         \includegraphics[width=0.2\linewidth]{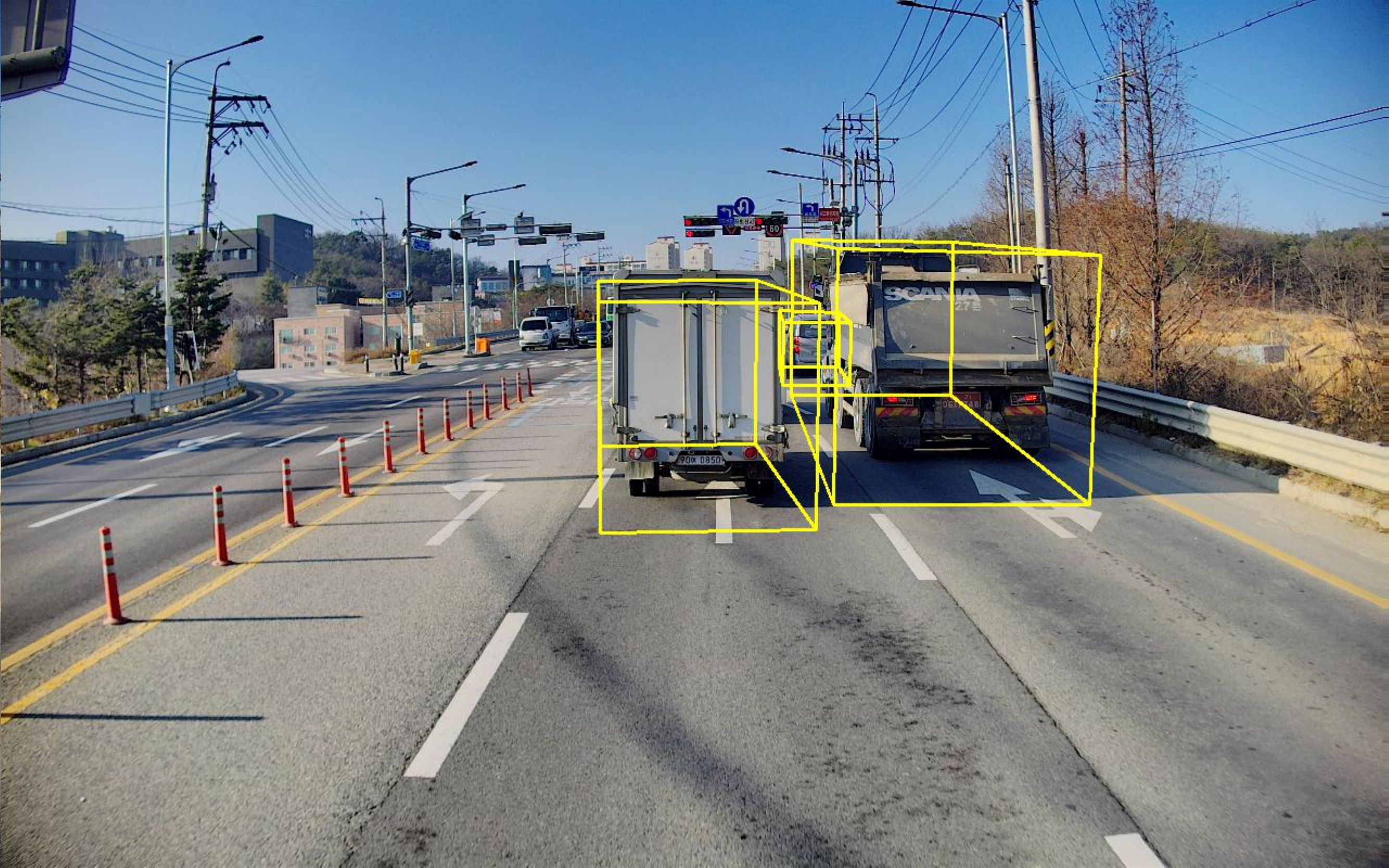} \\
         MonoDLE & Ours & MonoDLE & Ours \\
    \end{tabular}
    \vspace{-0.1cm}
    \caption{Qualitative results of the baseline model (MonoDLE \cite{ma2021delving}) and our method on real-world datasets. Training datasets are captured with a passenger car and test datasets are captured with a truck. We visualize the regressed 3D bounding box from MonoDLE and ours.}
    \vspace{-4mm}
\label{real_img}
\end{figure*}

\begin{figure}[t]
\scriptsize
    \centering
    \begin{tabular}{c@{\hspace{0.1mm}}} \includegraphics[width=7cm]{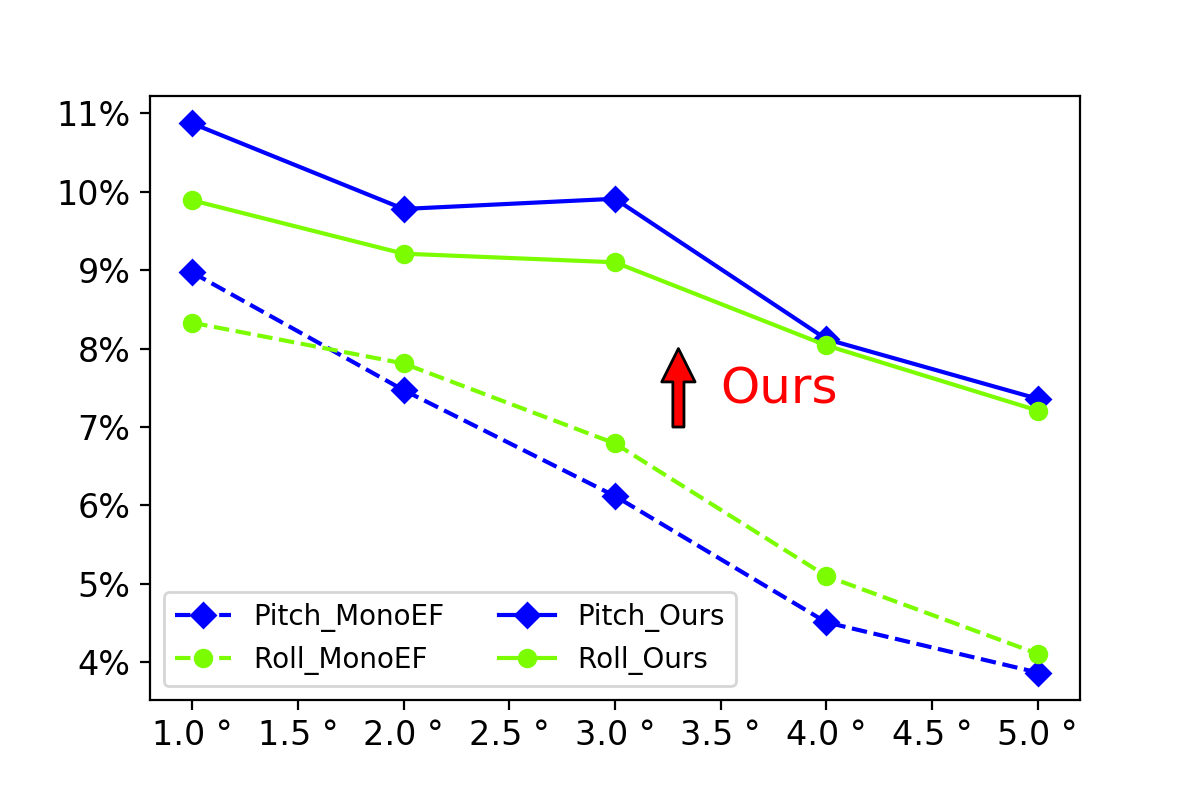}\\
    \end{tabular}
    \vspace{-0.1cm}
    \caption{Results comparison of MonoEF \cite{zhou2021monocular} and ours using $AP_{40}~(IoU=0.7)$.}
    \vspace{-0.65cm}
\label{monoef}
\end{figure}


\subsubsection{Quantitative/qualitative results from synthetic datasets}
\begin{figure}[t!]
\resizebox{\columnwidth}{!}{
    \centering
    \begin{tabular}{c@{\hspace{1mm}}c@{\hspace{1mm}}}
         \includegraphics[width=0.46\linewidth, height=2cm]{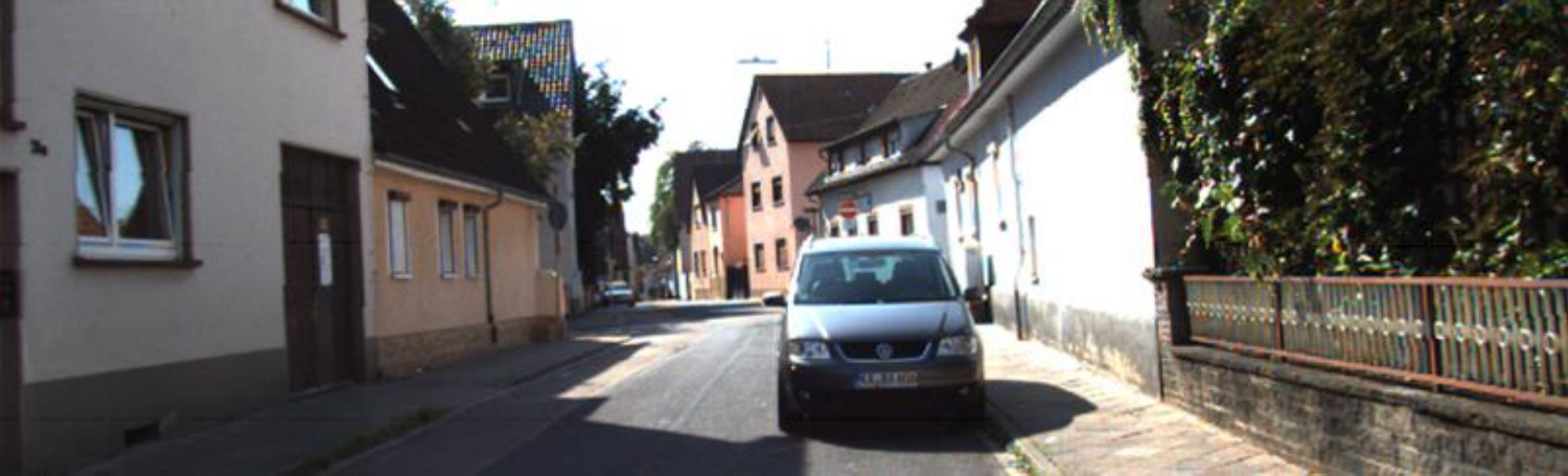}&
         \includegraphics[width=0.46\linewidth, height=2cm]{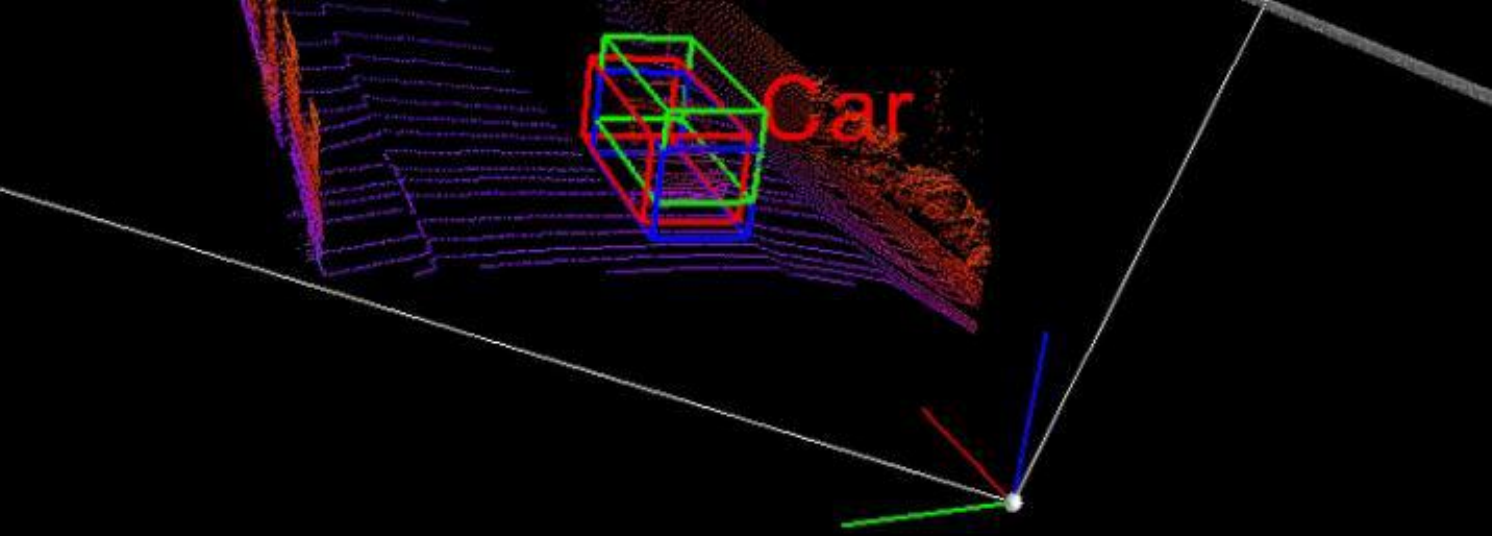}\\
         \includegraphics[width=0.46\linewidth, height=2cm]{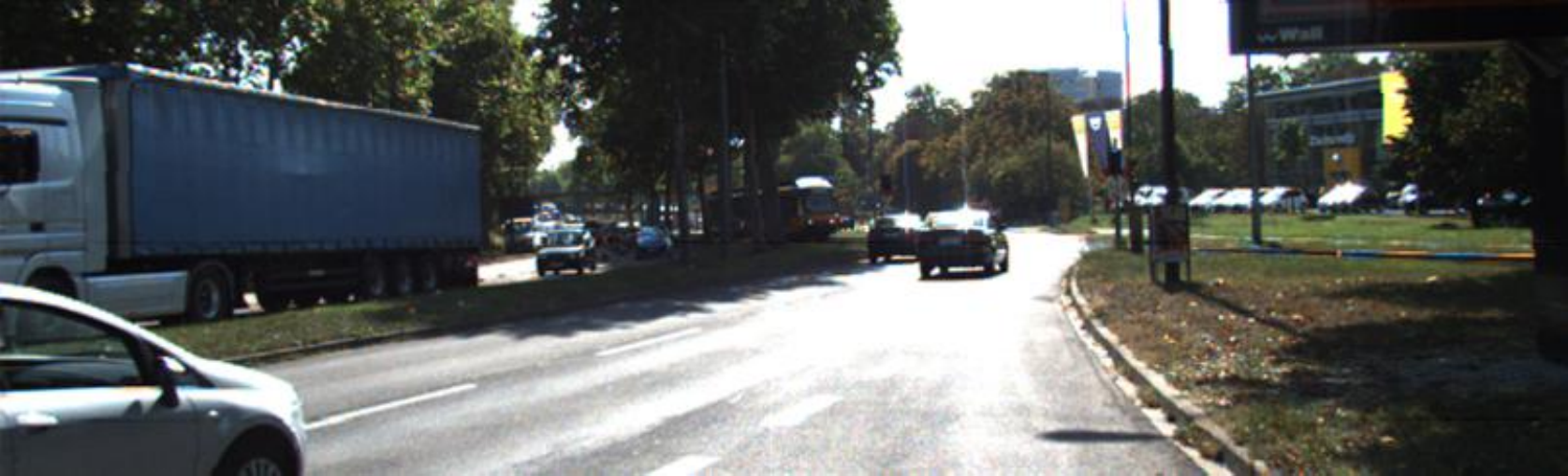}&
         \includegraphics[width=0.46\linewidth, height=2cm]{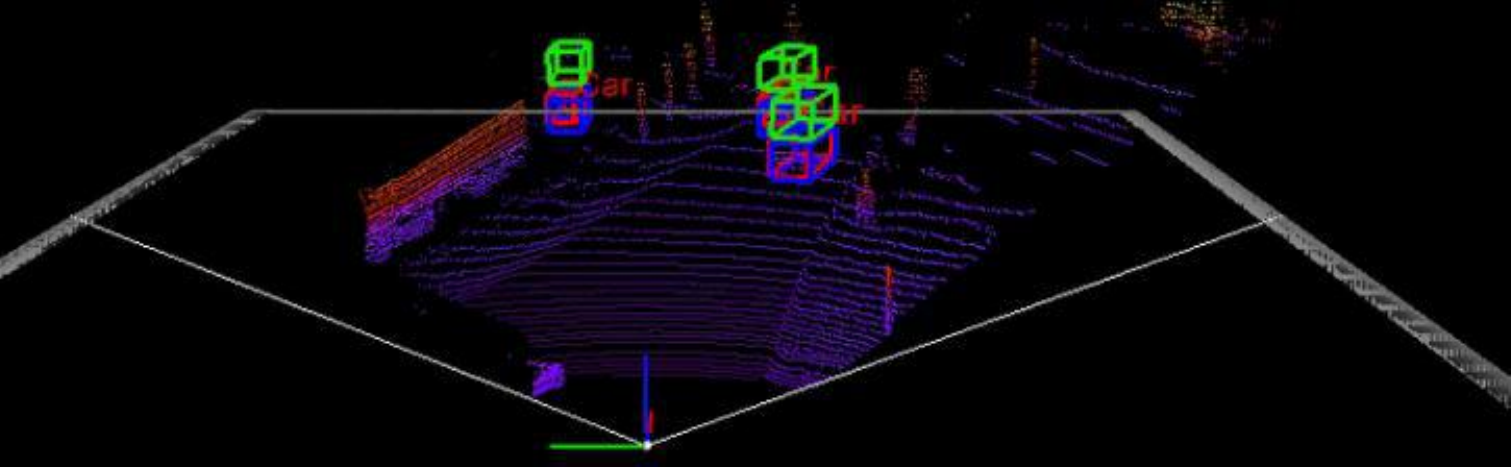}\\
         (a) Images with pitch rotation & (b) 3D bounding boxes in 3D \\
    \end{tabular}%
    }
    \caption{Visualization of 3D bounding box in 3D space. We visualize the 3D bounding boxes estimated from ours and the baseline model. Boxes marked in red, green and blue are the ground truth, baseline method, and our method, respectively.}
    \vspace{-0.8cm}
\label{visualize_3D}
\end{figure}

We synthesize the images by changing the roll ($r_{roll}$), pitch ($r_{pitch}$), yaw ($r_{yaw}$), and pitch with $y$-axis translation from $0^{\circ}$ to $5^{\circ}$ by increasing every $1^{\circ}$ for rotation changes. 
We evaluate the performance changes as the degree of rotation increases. 
As shown in \figref{3D_results_graph}, the average precision $AP_{3D}$ of all the methods, MonoGRNet, SMOKE, MonoDLE, and MonoFLEX decreases sharply as the degree of rotation increases on the roll and pitch axes. After $3^{\circ}$ rotation, the conventional methods show around $10\%$ of the original results and reach about $1\%$ with $5^{\circ}$ rotation.
Meanwhile, the proposed module avoids the performance degradation exhibited by all the competitive methods. 
It achieves about $80\%$ of the original results with $3^{\circ}$ rotation and about $60\%$ with $5^{\circ}$ rotation. 
We also observe that the performance from pitch rotated images and that from both pitch rotated and $y$-axis translated images are similar. This means the translation in the $y$-axis rarely results in performance drop.
We report the $AP_{3D}$ performances of all the competitive methods with ours in \tabref{Part_3D_results}. 
Some part of results are included in \figref{3D_results_graph}.
We additionally describe the results of yaw rotation ($r_{yaw}$) and $x$-axis translation.
The results from yaw rotation show that the conventional methods produce around $65\%$ of the original results.
The performance drop ratio is relatively lower than the results from roll or pitch rotation ($r_{pitch}, r_{roll}$) as shown in \figref{3D_results_graph}. 
This is because the monocular 3D object detection model learns the vehicle directly in the yaw axis while the heading directions in the roll and pitch axes have not been trained.
The compensation for object heading direction in the yaw axis does not significantly increase performance. 
The proposed module increases  performance by about $10\%$ of the conventional methods. 
We also visualize the 3D bounding boxes in 3D space in \figref{visualize_3D}.
The results show that our 3D object detection method outperforms the conventional methods not only in 2D projected space but also in real 3D space.
For the translation changes in $x$-axis or $y$-axis, the performance is almost retained, as shown in \tabref{Part_3D_results}.


Lastly, we compare our method to MonoEF \cite{zhou2021monocular}, which mitigates the extrinsic parameter perturbations of the 3D
detection task. 
The method predicts the camera extrinsic with respect to the road plane, then the feature maps of the input image are transferred using the estimated camera extrinsic.
Since the code for MonoEF is not released, we implement the algorithm without the extrinsic estimation part. 
For a fair comparison, we use the GT camera extrinsic and transfer the image features using GT extrinsic parameters.
As shown in \figref{monoef}, the proposed method outperforms the conventional compensation method.
This means the compensation of the 3D bounding box regression is more effective in an output space rather than the feature space. 

\subsubsection{Qualitative results from Real-world datasets}
To show the effectiveness of the proposed method, we conduct qualitative experiments with real-world datasets.
The training datasets are captured using a normal passenger car and the test datasets are captured using a truck. We use the same camera, which means the camera intrinsic is the same. 
The height of the camera of the two vehicles from the road is different.
To set a similar field of road view, the camera in the truck is tilted on the pitch axis.
The camera setting is equivalent to the TransY+Pitch in \figref{3D_results_graph} and \tabref{Part_3D_results}.
We train the baseline model, MonoDLE, with our real-world datasets in a supervised manner. 
We compare the results from the MonoDLE and ours in \figref{real_img}.
Our method does not require an additional training process. We use the MonoDLE model and the proposed compensation module.
The results show that with the existing method the estimated 3D bounding box is misaligned with the ground plane. On the other hand, our method estimates the 3D bounding box to fit the ground plane and the orientation of the object.

\subsection{Analysis of 2D object detection}
\begin{figure}[t!]
\scriptsize
    \centering
    \begin{tabular}{c@{\hspace{0.1mm}}} \includegraphics[width=9cm]{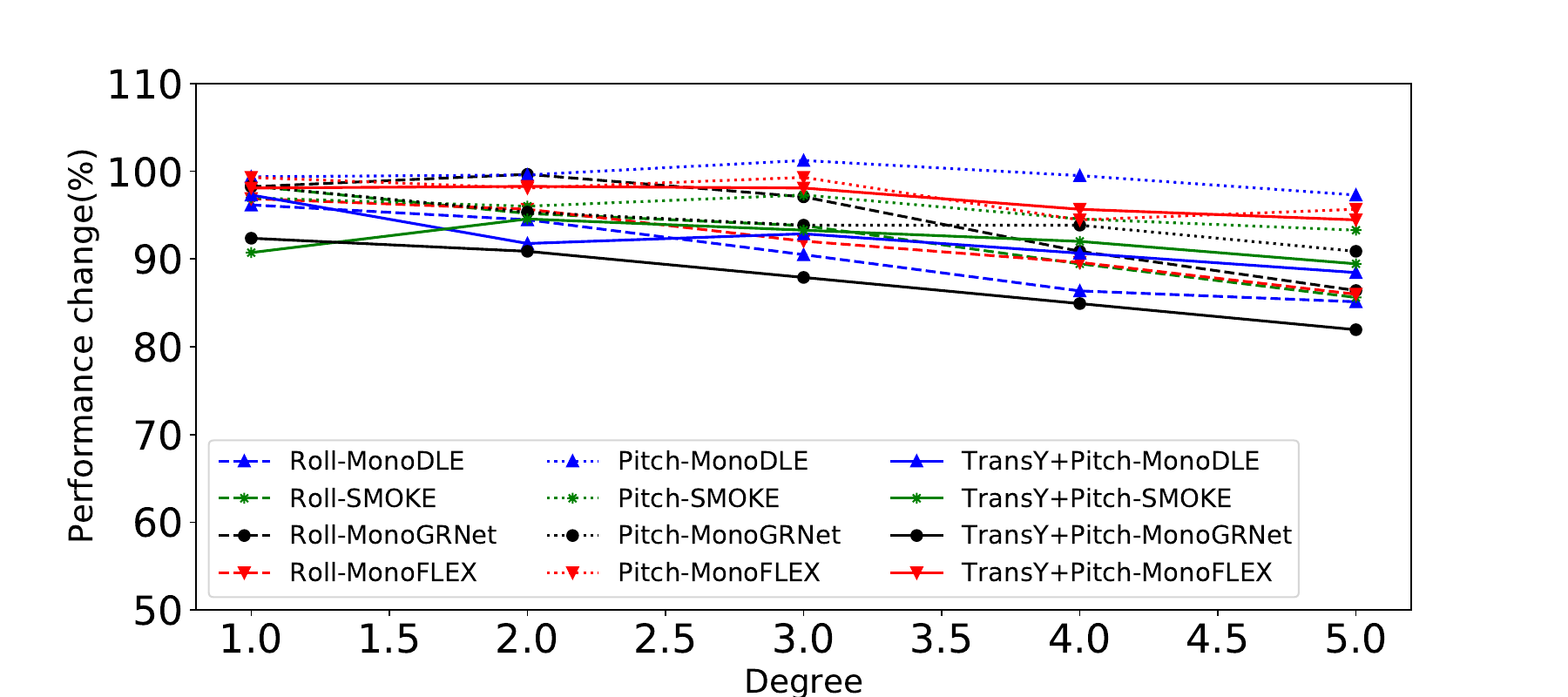}\\
    \end{tabular}
    \vspace{-0.1cm}
    \caption{Quantitative results on 2D object detection. Performance change ($\%$) means the percentage of the moderate 2D performance compared to original moderate 2D performance of each model (MonoGRNet \cite{qin2019monogrnet}, SMOKE \cite{liu2020smoke}, MonoDLE \cite{ma2021delving}, MonoFLEX \cite{zhang2021objects}).}
    \vspace{-2mm}
\label{results_2D}
\end{figure}

2D object detection is a sub-part of conventional 3D object detection networks.
2D detection is utilized as a guideline to regress projected 3D points.
We investigate the 2D detection performance of the networks with respect to the translation and rotation changes.
As shown in \figref{results_2D}, we observe that both translation and rotation changes rarely affect the performance of 2D object detection. 
Even with the $5^{\circ}$ rotated images, the 2D detection performance of the networks does not decrease significantly and is maintained at more than 80\% of the original performance. 

\subsection{Analysis of individual factors in 3D object detection}
We deeply analyze the individual prediction of 3D object detection networks.
We perform extensive experiments to figure out which of the various factors in the3D bounding box regression is significantly affected by small prediction errors or changes in camera rotation in \tabref{change_gt}. 
We use MonoDLE \cite{ma2021delving} as our baseline model for the ablation studies.
As described in \secref{sec:ours}, MonoDLE independently predicts each of the components, projected 3D points $[x_{2d}, y_{2d}]$, depth $z_{3d}$, yaw angle $r_{yaw}$, Bounding Box (BB) size $\mathbf{S}=[h,w,l]$, then the object 3D bounding box is computed using \eqnref{our_method}.
We measure the performance with $3^{\circ}$ pitch, yaw, roll rotated images, and original images.
We use the prediction values from MonoDLE, but replace the factor with GT values in \tabref{change_gt}-(a-e).
We use all GT values, but replace the factor with prediction values in \tabref{change_gt}-(f-j).
The results from the baseline model in the top row of \tabref{change_gt} demonstrate that the precision from pitch and roll rotated images is significantly degraded while the yaw rotation only slightly reduced performance. 
This means that camera rotations in the pitch and roll axis (not the yaw axis) are dominant factors affecting the drop in 3D object detection performance, just as was observed with \tabref{sec:state}.

\subsubsection{Bounding box size $\mathbf{S}=[h,w,l]$}
\label{sec:bb}
As reported in \tabref{change_gt}-(a), the performance of the baseline model, where the predicted BB size $\mathbf{S}=[h,w,l]$ is replaced by the GT value, marginally improved from 11.3\% to 13.3\%.
This means that the baseline network regresses very accurately on the 3D BB size, and small errors in the 3D BB prediction only slightly affect precision.
Moreover, we observe that the 3D object detection performance from the images rotated in pitch, yaw, and roll axis is also marginally improved (\textit{e.g.} $\text{pitch}_{3}$: 1.76\% to 2.12\%). Even though the GT BB size is utilized, the results with this level of precision are not available to be utilized. 
We can conclude that BB size is not a dominant factor in thedegraded performance at different camera settings.
The results in \tabref{change_gt}-(f), which use all GT values except the predicted BB size, show about $76\%$-$79\%$ $AP_{40}$ even with rotated images.
This is much higher precision than the results from the baseline.
This demonstrates that the 3D object detector accurately regresses BB size despite the rotated image being input.

\begin{table}[t!]
\resizebox{\columnwidth}{!}{
\centering
\begin{tabular}{lcccc}
\hline
\multicolumn{1}{c|}{replace factors with gt values} & \multicolumn{1}{c|}{original} & \multicolumn{1}{c|}{$pitch_3$} & \multicolumn{1}{c|}{$yaw_3$} & $roll_3$ \\ \hline
\multicolumn{1}{c}{Baseline (MonoDLE)} & \multicolumn{1}{c}{11.3} & \multicolumn{1}{c}{1.76} & 8.61 & 0.99 \\ \hline
(a) with   BB size                            & 13.3                          & 2.12                        & 9.76  & 1.38 \\
(b) with   projected3D                       & 12.0                          & 2.01                        & 9.42   & 1.27\\
(c) with   yaw angle                        & 11.8                          & 1.78                        & 10.5 & 1.31\\ 
(d) with   3D location                       & 77.6                          & 75.4                        & 74.8   & 72.1\\
(e) with   depth                             & 67.0                          & 36.7                        & 58.1  & 12.1\\\hline
(f) without   BB size                            & 78.2                          & 76.1                        & 77.4  & 78.9 \\
(g) without   projected3D                        & 76.4                          & 56.8                        & 74.4 & 18.9\\
(h) without   yaw angle                         & 69.8                          & 68.4                        & 69.4  & 67.1 \\ 
(i) without   3D location                        & 13.5                          & 11.8                        & 12.7 & 12.1 \\
(j) without   depth& 14.2& 12.5& 13.7 & 12.9 \\\hline
\end{tabular}
}
\caption{Performance analysis to investigate the dominant factors affecting the 3D detection performance. We use $AP_{40}~(IoU=0.7)$ under moderate setting on the KITTI validation set for 3D detection performance evaluation. (a)-(e) We replace the predictions to GT values. (f)-(j) We use all GT values and replace GT values to predictions.}
\label{change_gt}
\vspace{-0.4cm}
\end{table}

\subsubsection{Projected 3D object center $[x_{2d}, y_{2d}]$} 
\label{sec:p3d}
Similar to \secref{sec:bb}, we conduct a performance analysis with/without GT projected 3D points $[x_{2d}, y_{2d}]$ in \tabref{change_gt}-(b, g), respectively.
The results show aspects similar to those observed in the BB size analysis.
As reported in \tabref{change_gt}-(b), the performance of the baseline model with GT projected 3D points marginally improved, from 11.3\% to 12.0\%.
This means that the projected 3D points are accurately predicted.
Moreover, we observe that the 3D object detection performance from the images rotated in pitch, yaw, and roll axis is also marginally improved (\textit{e.g.} $\text{pitch}_{3}$: 1.76\% to 2.01\%) although the GT projected 3D points are utilized.
We can conclude that projected 3D points are not a dominant factor in  performance degradation with different camera settings.


\subsubsection{Heading direction, yaw angle $r_{yaw}$}
\tabref{change_gt}-(c, h) shows the performance with/without GT heading direction, yaw angle $r_{yaw}$.
These results also show aspects similar to those observed in the analysis of BB size and 3D projected center, as described in \secref{sec:bb} and \secref{sec:p3d}.
The results in \tabref{change_gt}-(c), applied GT yaw angle are almost similar to the results from baseline.
This means the predicted yaw angle is quite accurate to regress 3D bounding boxes.
The precision from yaw rotated image is much higher than that from roll or pitch rotated images (10.5\% vs 1.78\%, 1.31\%).
This means the conventional methods only consider an objects' yaw angle estimation.
The conventional method requires estimating all heading directions of vehicles, roll, pitch, and yaw for better regression.

\subsubsection{3D object center $[X_c,Y_c,Z_c]$ $\&$ depth $z_{3d}$}
\tabref{change_gt}-(d, e) shows the performance of the baseline model where the predicted 3D location $[X_c,Y_c,Z_c]$ or depth $z_{3d}$ is replaced by the GT values.
Surprisingly, all the results in \tabref{change_gt}-(d) including the original, $\text{pitch}_3$, $\text{yaw}_3$, and $\text{roll}_3$ achieve an average precision of over 70\%. 
This means the predictions of BB size and yaw angles are quite accurate while the 3D location prediction is relatively less accurate. 
With better estimates of 3D location, the performance of the 3D object detector will be drastically improved.
Even with the rotated images, the performance is retained. 
Although the 3D heading directions of the roll and pitch rotated cameras have some errors, the accuracy of the 3D positions mitigate the AP3D performance degradation.
In the conventional methods, the 3D object center is computed using the back-projection of the 2D projected center with the depth $z_{3d}$ by following \eqnref{eq:objectcenter}.
We additionally analyze  performance changes with GT depth value in \tabref{change_gt}-(e). 
The overall precision is lower than the results in \tabref{change_gt}-(d), but all results from original and rotated images achieves better performance than the baseline model.
Therefore, the depth and 3D location are the dominant causes of the performance drop. 

\section{Conclusion}
In this paper, we deeply analyze the factors that lead to performance degradation when pretrained 3D objection models are applied to other camera systems. We found that the camera pose changes, especially the roll and pitch rotation changes, critically affect the performance of the 3D object detection. 
Based on these observations, we propose a generalized 3D object detection method. 
Although the method is trained on just one specific camera setup, it is applicable to various camera systems.
The proposed module is generally applied to the recent monocular 3D object detectors, such as MonoGRNet \cite{qin2019monogrnet}, SMOKE \cite{liu2020smoke}, MonoDLE \cite{ma2021delving}, and MonoFLEX \cite{zhang2021objects}. 
Without any further training, the proposed method provides  about 6-to-10 times improved $AP_{3D}$ compared with state-of-the-art methods. 




{\small
\bibliographystyle{ieee_fullname}
\bibliography{egbib}
}

\end{document}